\documentclass{article}
\usepackage[utf8]{inputenc}
\usepackage{amsmath, amssymb, amsthm}
\usepackage{graphicx}
\usepackage{hyperref}
\usepackage{geometry}
\usepackage{tikz}
\usepackage{booktabs}
\usepackage{tabularx}
\usepackage{array}
\usepackage{adjustbox}
\usepackage{authblk}
\usepackage{subcaption}
\usepackage[table]{xcolor}
\usepackage{rotating}

\definecolor{green}{HTML}{01B401}
\definecolor{blue}{HTML}{0264FF}

\usetikzlibrary{positioning, calc, shapes.geometric}
\usepackage[backend=biber,style=numeric,sorting=none,natbib=true]{biblatex}
\DeclareMathOperator{\GBW}{\text{GBW}}
\DeclareMathOperator{\SR}{\text{SR}}

\usepackage{siunitx}
\title{Dense Associative Memories with Analog Circuits}
\author[1]{Marc Gong Bacvanski}
\author[2]{Xincheng You}
\author[3]{John Hopfield}
\author[4]{Dmitry Krotov}

\affil[1]{MIT}
\affil[2]{Independent Researcher}
\affil[3]{Princeton University}
\affil[4]{IBM Research}

\date{December 16 2025}

\begin{document}
\maketitle

\textbf{Abstract:} The increasing computational demands of modern AI systems have exposed fundamental limitations of digital hardware, driving interest in alternative paradigms for efficient large-scale inference. Dense Associative Memory (DenseAM) is a family of models that offers a flexible framework for representing many contemporary neural architectures, such as transformers and diffusion models, by casting them as dynamical systems evolving on an energy landscape. In this work, we propose a general method for building analog accelerators for DenseAMs and implementing them using electronic RC circuits, crossbar arrays, and amplifiers. We find that our analog DenseAM hardware performs inference in {\em constant} time independent of model size. This result highlights an asymptotic advantage of analog DenseAMs over digital numerical solvers that scale at least linearly with the model size. We consider three settings of progressively increasing complexity: XOR, the Hamming (7,4) code, and a simple language model defined on binary variables. We propose analog implementations of these three models and analyze the scaling of inference time, energy consumption, and hardware. Finally, we estimate lower bounds on the achievable time constants imposed by amplifier specifications, suggesting that even conservative existing analog technology can enable inference times on the order of tens to hundreds of nanoseconds. By harnessing the intrinsic parallelism and continuous-time operation of analog circuits, our DenseAM-based accelerator design offers a new avenue for fast and scalable AI hardware.

\section{Introduction}
{
\renewcommand{\thefootnote}{}
\footnotetext{Code available at \url{https://github.com/mbacvanski/AnalogET}.}
}

The unprecedented growth of artificial intelligence (AI) has driven demand for increasingly large and powerful models. At present, the field of generative AI is primarily driven by two settings: autoregressive transformers~\cite{vaswani2017attention} and diffusion models~\cite{sohl2015deep}. While these settings have demonstrated remarkable capabilities, they do so at a substantial computational cost. Their current implementations utilize digital computation, which faces fundamental challenges in energy efficiency, scalability, and latency, especially as model sizes and deployment demands continue to grow~\cite{jouppi2017datacenter, masanet2020recalibrating, patterson2021carbon}. These limitations have prompted interest in alternative computational paradigms that can efficiently handle the demands of modern AI workloads \cite{aifer2025solving}.  

Dense Associative Memories (DenseAMs) \cite{krotov2016dense,krotov2018adversarial}, a promising class of AI models which generalize Hopfield networks \cite{hopfield1982neural},  offer a new angle for tackling these problems. 
Unlike conventional feed-forward models, DenseAM inference can be defined through the temporal evolution of a state vector that is governed by a system of differential equations \cite{krotov2020large}. The state vector can be thought of as a particle exploring the surface of a high-dimensional energy landscape, which is the Lyapunov function of these dynamical equations. DenseAMs have been demonstrated to be flexible and expressive computational frameworks, capable of representing many primitives of modern AI architectures, such as attention mechanism \cite{ramsauer2020hopfield}, transformers \cite{hoover2024energy}, and diffusion models~\cite{hoover2023memory, ambrogioni2023search, pham2025memorization}. Furthermore, DenseAMs are error-correcting systems \cite{krotov2025modern}, a property ensuring that small perturbations of the desired temporal evolution of the state vector are corrected away by the dynamics of the network itself, rather than accumulated in time. Finally, DenseAMs are asymptotically stable---during the course of temporal evolution the computation happens during a finite transient period of time, which is followed by a steady state of neural activities. This asymptotic stabilization of dynamical trajectories removes the requirement to read out the ``answer'' to the computation problem at a precise moment of time, making DenseAMs robust to several classes of hardware imperfections. The confluence of the above properties makes DenseAMs appealing networks for analog hardware implementations that, on the one hand, are grounded in the physics of stable error-correcting dynamical systems and, on the other hand, are capable of representing computation in state-of-the-art AI networks.

In 1989, Hopfield argued that analog neural hardware can exceed the efficiency of digital implementations when the device physics directly instantiate the computational dynamics of the model itself~\cite{hopfield1990effectiveness}. Here, we revisit this idea with DenseAM models: we propose an analog circuit-based hardware accelerator design whose dynamics directly realize those of the DenseAM. We find that analog DenseAM hardware enables constant-time inference independent of model size, which is in stark contrast to GPU solvers and digital implementations. This intrinsic property makes DenseAM a natural fit for analog AI accelerators, and it highlights our circuit architecture as a viable hardware path to realize them. Using component specifications already demonstrated in fabricated devices, analog DenseAM hardware may achieve inference times on the order of tens to hundreds of nanoseconds, several orders of magnitude faster than digital systems.

By leveraging the natural dynamics of analog systems, this work establishes a new design of fast and scalable AI accelerators. 
The framework of DenseAMs and their efficient analog hardware implementations suggest a pathway for fundamentally redesigning the hardware-software interface for AI, enabling a new paradigm for fast, energy-efficient, and scalable computation.

\section{Dense Associative Memory basics}
\begin{figure}[t]
    \centering
    \includegraphics[width=\linewidth]{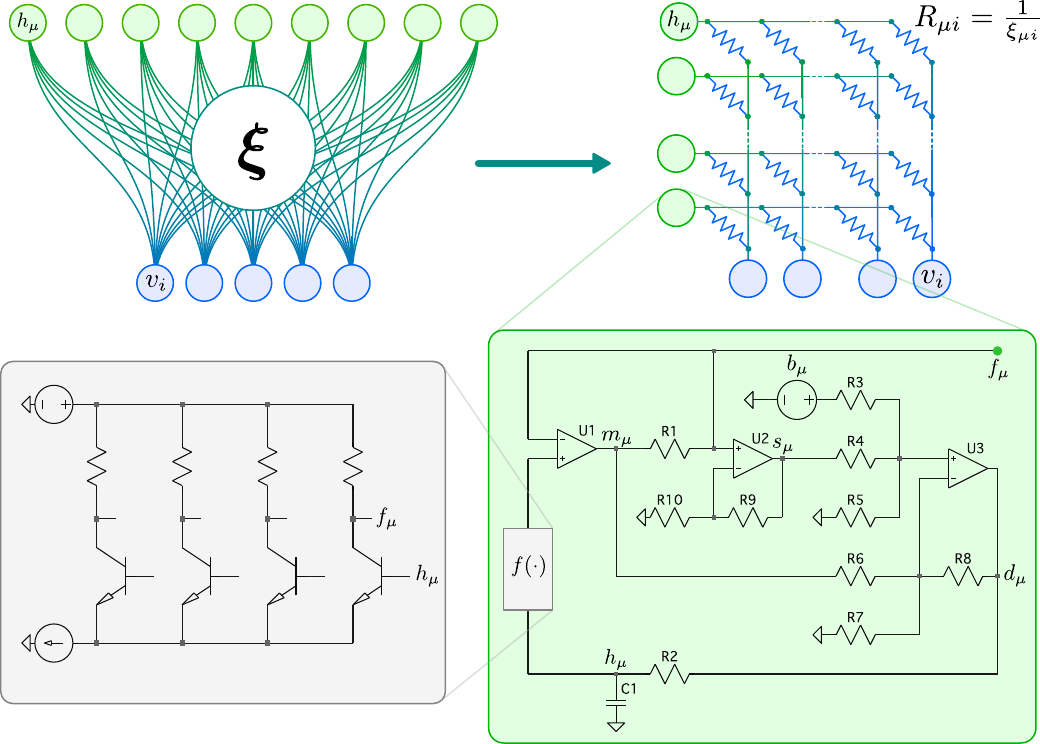}
    \caption{
    \textbf{Top left:} Bipartite neural network formulation, where hidden neurons $h_\mu$ and visible neurons $v_i$ are connected via symmetric synaptic weights $\boldsymbol \xi$. \textbf{Top right:} Circuit realization of symmetric weight matrix via resistive crossbar array. Each crosspoint encodes a weight $\xi_{\mu i}$ by its resistance $R_{\mu i}=1/\xi_{\mu i}$. \textbf{Lower right:} Circuit schematic of a single hidden neuron. It drives its row of the crossbar array with a voltage according to its activation $f_\mu$, and its internal dynamics are driven by the incoming current flowing into it from the crossbar array. \textbf{Lower left:} Softmax activation function built from bipolar junction transistors (some components not shown).
    }
    \label{fig:full_circuit}
\end{figure}

The DenseAM framework~\cite{krotov2020large, krotov2021hierarchical} provides a model that has straightforward neuronal dynamics, yet is surprisingly expressive in its ability to represent AI models including transformer attention, diffusion models, and associative memories. In its simplest form it is defined by two sets of neurons (typically called visible and hidden neurons) and a system of coupled non-linear differential equations governing their behavior, see \autoref{fig:full_circuit}. The visible neurons are characterized by their internal states $v_i$ and their outputs $g_i$, index $i=1\dots N_v$; while the hidden neurons have internal states $h_\mu$ and outputs $f_\mu$, index $\mu=1\dots N_h$. From the AI perspective, one can think about internal state of the neuron as a pre-activation of that neuron, and the output as a post-activation, which is obtained by applying an activation function to the pre-activation. From the biological perspective, one can think about the internal state of the neuron as a membrane voltage potential, and the output of that neuron as an axonal output, or a firing rate of that neuron. This framework admits both neuron-wise activation functions ($g_i = g(v_i)$, where $g(\cdot)$ is some continuous function, e.g., a ReLU), and collective activation functions such as softmax or layer normalization, which depend on the states of multiple neurons.  

The network parameters are stored in the synaptic weights $\boldsymbol{\xi}\in \mathbb{R}^{N_h\times N_v}$, whose matrix elements denoted by $\xi_{\mu i}$ can be either hand-engineered or learned. The time decay constants for the two groups of neurons are $\tau_v$ and $\tau_h$. With these conventions, the temporal evolution of the two groups of neurons can be expressed as 
\begin{equation}\label{eq:original differential equations}
\left\{
\begin{aligned}
\tau_v \frac{dv_i}{dt} &= \sum_{\mu=1}^{N_h} \xi_{\mu i} f_\mu + a_i - v_i\\
\tau_h \frac{dh_\mu}{dt} &= \sum_{i=1}^{N_v} \xi_{\mu i} g_i +b_\mu - h_\mu
\end{aligned}
\right.
\end{equation}
This forms a bipartite graph of neuronal connections, where the state of the hidden neurons is updated by the state of the visible neurons, and vice versa. 
Importantly, the same matrix $\boldsymbol{\xi}$ appears in both equations, once as $\boldsymbol{\xi}$ and again as $\boldsymbol{\xi}^\top$. Although this is sometimes described as using ``symmetric" weights, $\boldsymbol{\xi}$ is not assumed to be symmetric in the linear-algebraic sense; it is simply the same matrix used in both directions.
Finally, $a_i$ and $b_\mu$ denote biases, which are additional weights of the system and whose values may be hard-coded or learned depending on the application. 

The most important aspect of this model is the existence of a global energy function (Lyapunov function) that describes neuronal dynamics. To demonstrate this, it is most convenient to use the Lagrangian formalism \cite{krotov2020large,krotov2021hierarchical,krotov2025modern}. Each set of neurons is defined through a Lagrangian function of their internal states. The activation functions are defined as partial derivatives of that Lagrangian with respect to internal states. The total energy is the sum of energies of each set of neurons, plus the interaction energy. The energy of each set of neurons is a Legendre transformation of the corresponding Lagrangian (plus the term proportional to the bias). Thus, the global energy of \autoref{eq:original differential equations} is given by 
\begin{equation}
    E = \underset{\textcolor{blue}{\text{energy of visible neurons}}}{\underbrace{\Big(\sum\limits_{i=1}^{N_v} g_i (v_i - a_i) - \mathcal{L}_v\Big)}}\ +\ \underset{\textcolor{green}{\text{energy of hidden neurons}}}{\underbrace{\Big(\sum\limits_{\mu=1}^{N_h} f_\mu (h_\mu - b_\mu) - \mathcal{L}_h\Big)}}\ -\  \underset{\text{interaction energy}}{\underbrace{\sum\limits_{\mu=1}^{N_h}\sum\limits_{i=1}^{N_v} f_\mu\ \! \xi_{\mu i}\ \! g_i}} \label{global energy function}
\end{equation}
where the activation functions are defined as partial derivatives of the Lagrangians 
\begin{equation*}
    \ g_i = \frac{\partial \mathcal{L}_v}{\partial v_i}, \quad f_\mu = \frac{\partial \mathcal{L}_h}{\partial h_\mu}
\end{equation*}
For convex Lagrangians this global energy decreases with time on the dynamical trajectories of \autoref{eq:original differential equations}. If, additionally, the activation functions (and corresponding Lagrangians) are chosen in such a way that this energy is bounded from below, the dynamical trajectories are guaranteed to arrive at a stable fixed point of activations. The dynamical equations typically have many asymptotic fixed  points, which correspond to local minima of the energy function in \autoref{global energy function}. Both properties above (convexity of Lagrangians and lower-bounded energy) are satisfied for all settings studied in this paper. By picking different nonlinear activation functions (or corresponding Lagrangians), this system yields a variety of models that can describe associative memory, softmax attention, and other commonly used settings in AI~\cite{krotov2020large,ramsauer2020hopfield,krotov2021hierarchical,tang2021remark,hoover2022universal}. 

A particularly relevant example for modern sequence modeling is the Energy Transformer (ET)~\cite{hoover2024energy}, which reformulates transformer's inference pass as a gradient flow on an energy function defined over the set of tokens. The ET block contains two contributions to the energy function: attention energy and the Hopfield network. The energy attention module routes the information between the tokens, while the Hopfield module aligns the tokens with the manifold of token embeddings.  In our implementation, the context tokens act as a set of dynamically instantiated memories that interact with the predicted token through a DenseAM-like energy. In \autoref{sec:denseam et lm} we exploit this connection to construct an Analog Energy Transformer (Analog ET) whose continuous-time dynamics are implemented directly in hardware using our DenseAM circuit primitives.

\section{Related work}
Early analog implementations of associative memories focused on the classical Hopfield network. Foundational designs, such as continuous-time analog circuits~\citep{hopfield1984neurons, tank1988simple} and later demonstrations using amorphous-silicon resistors~\citep{graf1986vlsi}, memristive devices~\citep{guo2015modeling, hu2015associative}, and phase-change memories~\citep{eryilmaz2014brain}, targeted the quadratic Hopfield energy function. These works emphasize device engineering and memory-cell design rather than system-level dynamics, and inherit the limited storage capacity and representational power of traditional Hopfield networks. That line of research is largely concerned with how to fabricate programmable resistance elements themselves; our work assumes programmable conductances as a given primitive and focuses on the continuous-time dynamics that operate on top of them. Our work also differs from these works by addressing DenseAMs with higher-order energy functions and continuous-valued states. 

Another direction is the use of cavity-QED systems for associative memory. \citet{marsh2021enhancing} analyze a confocal cavity implementation of a quadratic Hopfield network and show that the cavity dynamics induce a descent-like relaxation rule on spin states. Their model remains restricted to quadratic energies and binary spins, and operates in a cryogenic, cavity-QED setting. Our work instead targets higher-order DenseAMs with continuous states, and emphasizes scalable, room-temperature analog microelectronics with explicit hardware-aware dynamical analysis.

More recent physical implementations move beyond purely quadratic energies. \citet{musa2025dense} propose a free-space optical realization of the higher-order DenseAM energy. Their system constructs a static physical representation of the energy landscape, but inference relies on an external digital controller that performs iterative spin-flip updates. Thus, the hardware computes energies, while the optimization dynamics remain digital. In contrast, our analog microelectronic architecture embeds the gradient flow itself into hardware: inference is performed by a single continuous-time evolution rather than by discrete digital updates.

\section{DenseAM circuit design}
Here, we introduce a novel architecture for a class of analog electronic hardware accelerators that models \autoref{eq:original differential equations}'s system of nonlinear differential equations using time evolution. Our DenseAM design shown in \autoref{fig:full_circuit} is comprised of two sets of neurons that interact through a resistive crossbar array. The resistive crossbar array turns voltage differences between neurons into currents flowing between the neurons according to synaptic weights, and each neuron's internal circuitry converts those currents into dynamics that reproduce \autoref{eq:original differential equations}.

\paragraph{Resistive weights as a crossbar array.}
The crossbar array construction is a canonical design of matrix-vector multiplication using analog electronics~\cite{hopfield1990effectiveness,mead2012analog}, and is a natural fit for the weight matrix $\boldsymbol\xi$ in our model. Traditionally, each crosspoint between a row and column line is connected by a resistor (often memristors, RRAM, or other analog memories that produce resistances), a vector of input voltages is applied at row lines, and the column lines are held at ground typically via a transimpedance amplifier. By Ohm's law, each resistive crosspoint produces a current that multiplies the row's input voltage by the inverse of the resistance. Because currents add along each column line, the total current output at a column is the inner product between the vector of input voltages and the column's conductance vector. Thus, the array as a whole implements a parallel analog matrix multiplication of the form $I_\text{out}=GV_\text{in}$, where $G$ is the matrix of conductances (inverse of resistances).

Unlike a traditional crossbar array whose rows are driven at a fixed voltage and whose columns are held at ground, our DenseAM circuit design uses each weight bidirectionally, exactly representing the bidirectional connections between visible and hidden neurons. As a result, the current flowing into each neuron corresponds to the weighted sum of the differences between visible and hidden neuron activations. For example, for hidden neuron $\mu$, this current is ${\sum_i \xi_{\mu i}(g_i-f_\mu)}$. This construction enables weight symmetry to be enforced by hardware sharing: both forward and reverse weights are realized by the same resistive elements. Importantly, as long as weights are represented as conductances, they must be non-negative.

\paragraph{Design of a single neuron.}
Each neuron in the circuit computes its dynamics by integrating the currents it receives from the crossbar array, which represent weighted differences between its own activation and those of connected neurons. Considering a hidden neuron (the design for visible neurons is symmetric by design), the neuron’s internal voltage $h_\mu$ is stored on capacitor \texttt{C1} and evolves in continuous time, while the neuron’s activation $f_\mu$ is obtained by passing $h_\mu$ through a nonlinear function (e.g.\ ReLU or softmax). 

The current flowing into hidden neuron $\mu$ is produced by its interaction with all visible neurons via the synaptic weights $\xi_{\mu i}$ for $i=1,\dots,N_v$. Specifically, this is as a weighted sum of the differences between neuron activations: ${\sum_i \xi_{\mu i}(g_i-f_\mu)}$. Inside each neuron, a ``self'' path scales $f_\mu$ to produce the voltage $s_\mu=f_\mu \sum_i \xi_{\mu i}$. This term is added to the value of the incoming current so that the $-f_\mu\sum_i \xi_{\mu i}$ term is cancelled inside each neuron. As a result, the hidden state, represented as the voltage across capacitor $C_1$, integrates only the desired weighted input plus any external stimulus $b_\mu$. Its dynamics reduce to the canonical DenseAM form with a time constant of $R_2C_1$:
\begin{equation}
    R_2 C_1 \frac{dh_\mu}{dt} = \sum_{i=1}^{N_v} \xi_{\mu i}\, g_i + b_\mu - h_\mu
\end{equation}
Elementwise (or vectorized) nonlinearities then produce activations $g_i=g(v_i)$ and $f_\mu=f(h_\mu)$ (e.g., ReLU, softmax) across the visible and hidden neurons. See Appendix~\ref{appendix:neuron-design} for the full circuit derivation.

\section{Analog DenseAM Examples}
We begin by studying two examples of the proposed design: the XOR task, and the (7,4) error-correcting Hamming code. 
\subsection{XOR}\label{sec:xor}

\begin{figure}
  \centering
  \begin{minipage}[t]{0.49\linewidth}
    \centering
    \includegraphics[width=\linewidth]{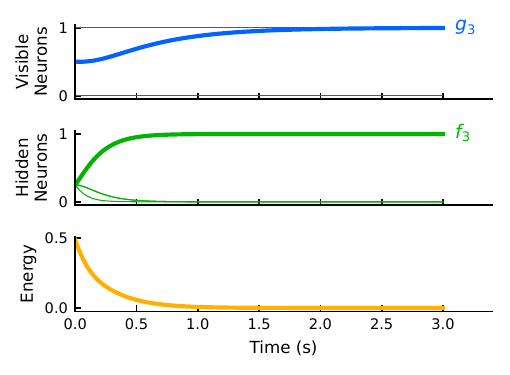}
    \caption{
    Solving XOR with a DenseAM. Visible neuron $g_3=v_3$ serves as the output, while the two input neurons (unlabeled, thin lines) are clamped at $1$ and $0$ for True and False. Output $v_3$ is initialized at $0.5$ and converges to a positive prediction of $1$. The activation of the hidden neuron $f_3$ for the truth-table row (1, 0, 1) becomes highly activated, with others (fine lines) are suppressed by softmax. Energy (\ref{global energy function}), or equivalently (\ref{eq: effective energy function}), decreases monotonically along the inference trajectory.
    }
    \label{fig:xor-inference}
    \end{minipage}\hfill
    \begin{minipage}[t]{0.49\linewidth}
    \centering
    \includegraphics[width=\linewidth]{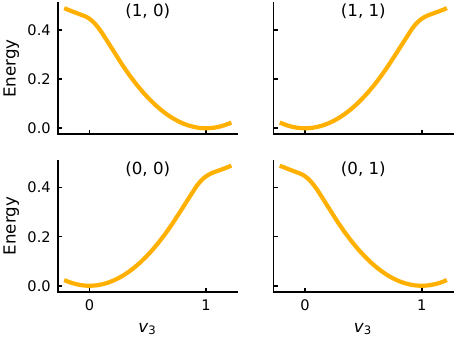}
    \caption{XOR energy landscape of neuron $v_3$ under different settings of visible input neurons $v_1$ and $v_2$. Minima in the energy function correspond to stationary points of the dynamics. Gradient flow dynamics bring $v_3$ to these attractor points, resulting in correct XOR outputs.}
    \label{fig:xor-energy-landscape}
  \end{minipage}
\end{figure}

The XOR problem is a canonical test for nonlinear representation and inference, as it cannot be solved by any linear model. We show a minimal DenseAM model for the XOR task, illustrating how energy-based dynamics can solve this simple task with a continuous-time analog system. The network consists of $N_v=3$ visible neurons, and $N_h=4$ hidden neurons. At $t=0$ visible neurons $v_1$ and $v_2$ are initialized at their input values corresponding to the input bits. The last visible neuron $v_3$ is initialized at $v_3 = 0.5$. The hidden neurons are initialized at zero. The two input visible neurons remain clamped during the dynamics, while the third output visible neuron and the hidden neurons evolve in time according to (\ref{eq:original differential equations}). Each row of the memory matrix $\boldsymbol \xi$ corresponds to a row of the XOR truth table. The visible neurons use an identity activation function where $g_i=v_i$, and the hidden neurons use a softmax activation. The biases are set as
\begin{align*}
    a_i=0,\quad b_\mu = -\frac{1}{2} \sum\limits_{i=1}^{N_v} \big(\xi_{\mu i}\big)^2
\end{align*}

\autoref{fig:xor-inference} shows the temporal evolution of visible and hidden neuron activations, as well as the total energy, during inference on the XOR input $(1, 0)$. The output visible neuron's activation $g_3$ gradually converges to the correct prediction of $1$, while the hidden neuron associated with that memory, $f_3$, becomes strongly activated and the remaining hidden neurons are suppressed by the softmax nonlinearity. The system's energy decreases monotonically throughout the trajectory and stabilizes once the network reaches its fixed-point prediction.
\autoref{fig:xor-energy-landscape} depicts the system's energy landscape as a function of output neuron $v_3$ for different clamped input configurations $(v_1, v_2)$. In each case, the energy exhibits a clear convex minimum at the correct XOR output, demonstrating that gradient flow along the energy surface drives $v_3$ reliably toward the correct prediction.
As shown in \autoref{sec:xor-spice}, we validate our circuit design and dynamics using SPICE simulation.

To analyze this DenseAM, it is instructive to consider the limit $\tau_h\rightarrow 0$. Since the second equation in (\ref{eq:original differential equations}) is linear in hidden units $h_\mu$, they can be integrated out. With $\sum_{\mu=1}^{N_h} f_\mu=1$, the resulting dynamics of the visible neurons can be written as
\begin{align}
    \tau_v \frac{dv_i}{dt} = \sum_{\mu=1}^{N_h} \big(\xi_{\mu i}  - v_i\big) f_\mu \quad \text{where} \quad f_\mu = \text{softmax}\Big( -\frac{\beta}{2} \sum\limits_{i=1}^{N_v} (\xi_{\mu i} - v_i)^2\Big)
    \label{eq:xor-integrated-visible-dynamics}
\end{align}
The effective energy on the visible neurons can be written as 
\begin{equation}
    E^{\text{eff}}(\boldsymbol{v})=-\frac1\beta\log\sum\limits_{\mu=1}^{N_h}\exp\Big[-\frac{\beta}{2}\sum\limits_{i=1}^{N_v} (\xi_{\mu i} - v_i)^2\Big] \label{eq: effective energy function}
\end{equation}
Intuitively, each hidden neuron computes a squared Euclidean distance between the visible state and its stored pattern $\boldsymbol{\xi}_{\mu}$. The softmax nonlinearity assigns higher weight to the pattern closest to the current state of the visible neurons. The resulting visible neuron dynamics are gradient flow for this effective energy.
It is important to note that memories in this implementation are represented by conductances of the crossbar array, which are always positive. For this reason, matrix elements of memories $\xi_{\mu i}$ must be positive, necessitating the use of the bias terms, which are just voltage sources that can be arbitrarily signed.

While a time constant of $\tau_h=0$ is impossible to physically construct due to finite conductances and nonzero capacitances, setting $\tau_h \ll \tau_v$ realizes the same adiabatic limit in practice. When hidden neurons evolve much faster than visible ones, they reach their steady state almost instantaneously for each configuration of visible neurons. The result is an adiabatic elimination of hidden dynamics, yielding the effective visible-only dynamics above. In practice, for the XOR task, even a relatively modest $\tau_h = \tau_v/10$ ratio yields perfect performance.

\subsection{Hamming (7,4) code}
\begin{figure*}[t]
\centering
\begin{minipage}[c]{0.48\textwidth}
  \centering
  \includegraphics[width=\linewidth]{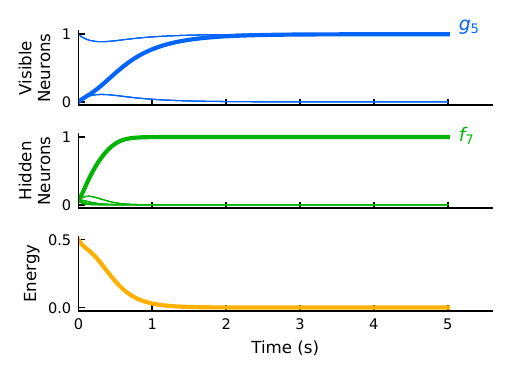}
  \captionsetup{type=figure}
  \captionof{figure}{Correcting a bit error in a Hamming (7,4) code. Visible neuron $g_5$ flips indicating the bit flip error happened on the $5$th codeword bit. $f_7$ is the hidden neuron corresponding to the memory of the correct codeword. Thin lines correspond to the other neuron activations.}
  \label{fig:hamming74-inference}
\end{minipage}
\hfill
\begin{minipage}[c]{0.48\textwidth}
  \centering
  \small
  \setlength{\tabcolsep}{6pt}
  \renewcommand{\arraystretch}{0.95}
\begin{tabular}{c|c} \toprule Data bits $(d_1 d_2 d_3 d_4)$ & Codeword $(c_1 c_2 c_3 c_4 c_5 c_6 c_7)$ \\ \midrule 0000 & 0000000 \\ 0001 & 0001111 \\ 0010 & 0010110 \\ 0011 & 0011001 \\ 0100 & 0100101 \\ 0101 & 0101010 \\ 0110 & 0110011 \\ 0111 & 0111100 \\ 1000 & 1000011 \\ 1001 & 1001100 \\ 1010 & 1010101 \\ 1011 & 1011010 \\ 1100 & 1100110 \\ 1101 & 1101001 \\ 1110 & 1110000 \\ 1111 & 1111111 \\ \bottomrule \end{tabular}  \captionsetup{type=table}
  \captionof{table}{Valid codewords of the Hamming(7,4) code, ordered by their 4-bit data payload.}
  \label{tab:hamming74-codewords}
\end{minipage}
\end{figure*}

The Hamming (7,4) code is an error-correcting code that encodes 4 data bits into a 7-bit codeword by adding 3 parity bits. The resulting 7-bit strings are special: only certain patterns are valid codewords, and they are spaced apart so that if a single bit is flipped, the error can be detected and corrected~\cite{hamming1950error}. Table~\ref{tab:hamming74-codewords} lists the 16 codewords corresponding to four arbitrary data bits.

Unlike the XOR case where the only evolving neuron is the readout bit, the Hamming (7,4) code may require flipping the value of any one of the visible neurons. During inference, the visible neurons are initialized to the corrupted 7-bit input word. All neurons are left free to evolve, and the dynamics relax the state toward the nearest stored codeword. Energy minima are located at the valid codewords, so the network converges to the correct code provided the error is within the Hamming radius of 1. Thus, the DenseAM replicates the standard decoding property of the Hamming (7,4) code: any single-bit flip is corrected automatically. \autoref{fig:hamming74-inference} illustrates a case where a flipped bit $g_5$ is restored during convergence.

The Hamming (7,4) model's 7 visible neurons, each corresponding to a codeword bit, are connected to 16 hidden neurons, each representing one valid codeword. The weight matrix $\boldsymbol\xi\in\{0,1\}^{16\times7}$ is formed by stacking the 16 codewords as its rows. Visible neurons have the identity activation, hidden neurons use a softmax activation, and biases are chosen as in the XOR case to give the same integrated-out visible dynamics as~\eqref{eq:xor-integrated-visible-dynamics}.

\section{Analog Energy Transformer (Analog ET) via DenseAM}\label{sec:denseam et lm}
\begin{figure}
    \centering
    \includegraphics[width=\linewidth]{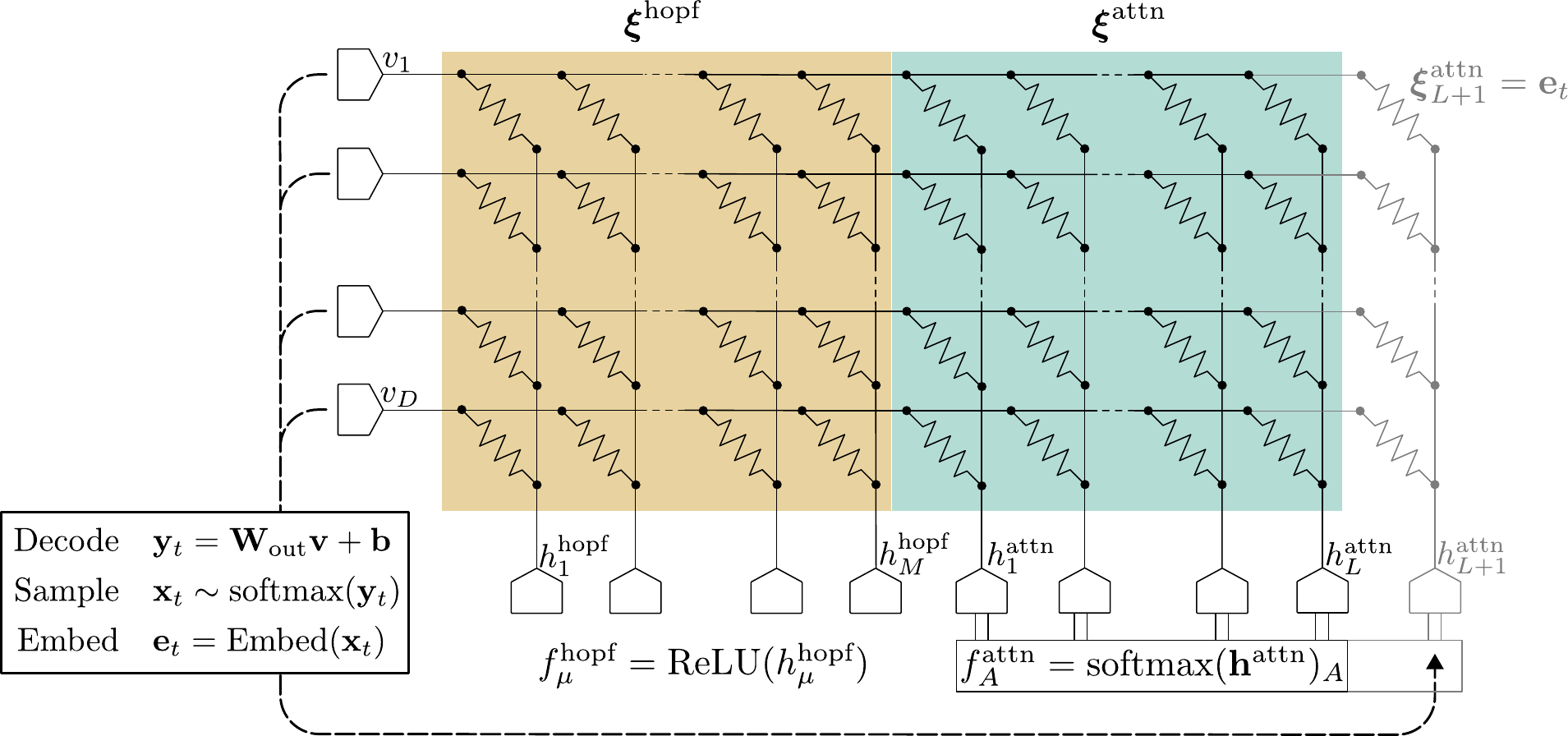}
    \caption{Analog ET circuit demonstrating the autoregressive inference procedure. A newly inferenced token is decoded, sampled, and re-embedded to obtain the weight vector $\boldsymbol\xi_{L+1}^\text{attn}$, which is set as the weight vector for a new hidden neuron $h_{L+1}^\text{attn}$ in the energy attention block (light gray on right). For this layout we have flipped the crossbar array, so that indices $A$ and $\mu$ run horizontally and index $i$ runs vertically.}
    \label{fig:et-circuit}
\end{figure}
Our DenseAM circuit construction can be used to build more complex energy-based models, such as the transformer-like architecture proposed in the Energy Transformer paper~\cite{hoover2024energy}. For causal next-token prediction with a single attention head, the Energy Transformer's energy function can be written as the following (See \autoref{appendix:et-energy-function} for full derivation):
\begin{align}
E
=\tfrac12\|\mathbf v-\mathbf a\|^2 &- \mathbf v^\top\left(\left(\boldsymbol{\xi^\text{attn}}\right)^\top \mathbf f^\text{attn} + \left(\boldsymbol{\xi^\text{hopf}}\right)^\top \mathbf f^\text{hopf}\right) +\left(\mathbf f^\text{attn}\right)^\top\left(\mathbf h^\text{attn}-\mathbf b\right)+\left(\mathbf f^\text{hopf}\right)^\top\left(\mathbf h^\text{hopf}-\mathbf c\right)\nonumber\\
&-L^\text{attn}\left(\mathbf h^\text{attn}\right)-L^\text{hopf}\left(\mathbf h^\text{hopf}\right)
\end{align}
with the activation functions and their Lagrangians defined as
\begin{alignat}{2}
    f^\text{attn}_A &= \text{softmax}(\beta \mathbf h^\text{attn})_A, &\quad
    L^\text{attn}(\mathbf h) &= \tfrac{1}{\beta}\log\sum_{A=1}^L e^{\beta h_A} \\
    f^\text{hopf}_\mu &= \text{ReLU}(h^\text{hopf}_\mu), &\quad
    L^\text{hopf}(\mathbf h) &= \tfrac{1}{2}\sum_{\mu=1}^M \Big[\text{ReLU}(h_\mu)\Big]^2
\end{alignat}
where $\mathbf a$, $\mathbf b$, and $\mathbf c$ correspond to the biases of the visible neurons, attention hidden neurons, and Hopfield network hidden neurons, respectively. The $L$ context tokens are indexed by $A$, and the $M$ hidden neurons of the Hopfield network are indexed by $\mu$.  Because the visible units use an identity activation function, $g_i=v_i$ using the languge of \autoref{eq:original differential equations}, the gradient flow of the energy yields the dynamics:
\begin{align}
    \tau_v \dot{\mathbf{v}} 
        &= -\frac{\partial E}{\partial \mathbf{v}}
           = \left(\boldsymbol{\xi}^{\text{attn}}\right)^\top \mathbf{f}^{\text{attn}}
             + \left(\boldsymbol{\xi}^{\text{hopf}}\right)^\top \mathbf{f}^{\text{hopf}} + \mathbf a - \mathbf v\\
    \tau_h \dot{\mathbf{h}}^{\text{attn}} 
        &= -\frac{\partial E}{\partial \mathbf{f}^{\text{attn}}}
           = \boldsymbol{\xi}^{\text{attn}} \mathbf{v} 
             + \mathbf{b} - \mathbf{h}^{\text{attn}}\\
    \tau_h \dot{\mathbf{h}}^{\text{hopf}} 
        &= -\frac{\partial E}{\partial \mathbf{f}^{\text{hopf}}}
           = \boldsymbol{\xi}^{\text{hopf}} \mathbf{v} 
             + \mathbf{c} - \mathbf{h}^{\text{hopf}}
\end{align}
In this formulation, $\mathbf v$ represents the embedding of the output (next) token, and its evolution is driven by two terms: one term from the energy attention with weights $\boldsymbol\xi^\text{attn}$ and hidden neuron activations $\mathbf f^\text{attn}$, and one term from the Hopfield network with weights $\boldsymbol\xi^\text{hopf}$ and hidden neuron activations $\mathbf f^\text{hopf}$. The weights of the energy attention DenseAM are dependent on the context: for a token dimension $D$, context length $L$, and the task of predicting the token at index $L+1$, the weights $\boldsymbol\xi^\text{attn}\in\mathbb R^{L\times D}$ are generated by embedding each token of the context via a learned embedding matrix applied to each context token. In contrast, the Hopfield network weights $\boldsymbol\xi^\text{hopf}$ are learned during training and fixed at inference. The number of memories in the Hopfield network is a hyperparameter $M$, such that $\boldsymbol\xi^\text{hopf}\in\mathbb R^{M\times D}$.

This system suggests a hardware implementation where $\mathbf v$ interacts with two independent DenseAMs, one for the energy attention and one for the Hopfield term, which can share the same physical crossbar structure. \autoref{fig:et-circuit} shows that the circuit structure remains a crossbar array (like Figure~\ref{fig:full_circuit}), but with two distinct classes of hidden neurons. Because of the summation of currents along each row of the crossbar array, the incoming current to visible neuron $v_i$ is the sum of contributions from the energy attention block and from the Hopfield network block. The energy attention hidden neurons $\mathbf h^\text{attn}$ use a softmax activation function, while the Hopfield network hidden neurons $\mathbf h^\text{hopf}$ use a ReLU activation.

\subsection{Analog Energy Transformer on the parity task}
We build and evaluate the Analog ET on the $L$-bit parity task, which can be thought of as an elementary ``language model'':  given bits $\text{bit}_1,\dots, \text{bit}_L$, predict $\text{bit}_{L+1}=\left(\sum_{A=1}^L\text{bit}_A\right)\mod{2}$. Parity is instructive because it requires a representation of a global, order-$L$ interaction, precluding linear and shallow models from representing it efficiently. A successful model must be able to form high-order interactions in order to generalize. We formulate parity as a next-token prediction problem: given an $L$-bit string as context, predict its parity in the next token.

We train the Analog ET model digitally using backpropagation through time~\cite{werbos2002backpropagation} implemented with Jax's automatic differentiation. The resulting weights can be deployed onto the analog hardware; in our experiments we simulate the dynamics of hardware with the Diffrax~\cite{kidger2021on} ODE solver library. On the 8-bit parity task, our model achieves 100\% accuracy on the hold-out validation set of 52 bit strings, demonstrating clear generalization capabilities. 
See Appendix \ref{appendix:parity-training} for more details on training and model design.

\begin{figure}
    \centering
    \includegraphics[width=1\linewidth]{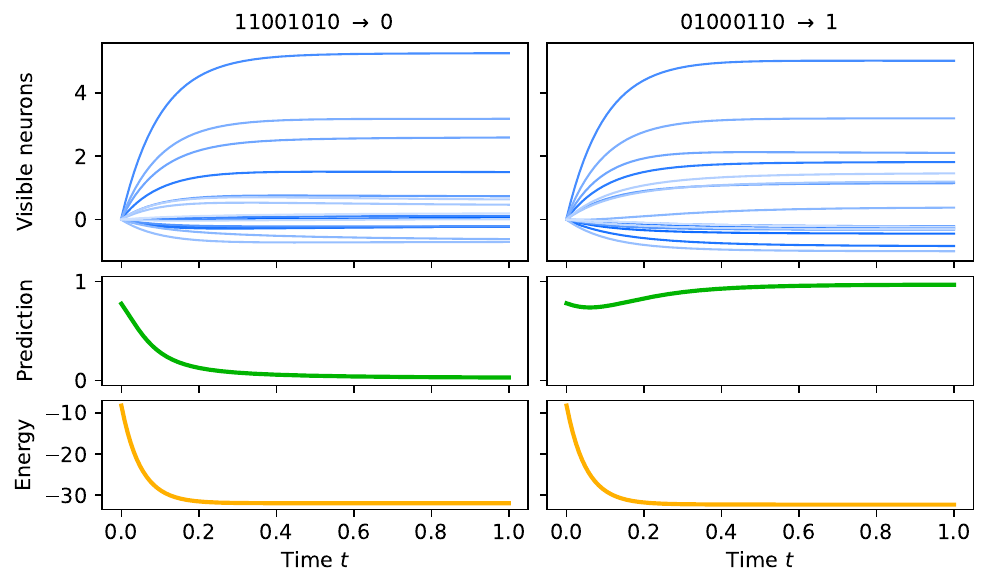}
    \caption{Inference of parity Analog ET on two example 8-bit strings. Top row plots the visible neurons $v_i$ over time, middle row plots the decoded token prediction, bottom row plots the energy that monotonically decreases during inference. After a transient period of computation, the network arrives at a steady-state, making the result of the computation robust against the precise timing of the readout.}
    \label{fig:parity-inference}
\end{figure}

\autoref{fig:parity-inference} shows the dynamics of the visible neurons and energy during two example inference runs of the Analog ET. Notably, the visible neuron values are constant by the end of the inference period, meaning that the inference remains highly stable to mismatch and delay in timing during readout. A single sample-and-hold and switching circuit would enable a single Analog-Digital Converter (ADC) to read out all the visible neurons at convergence, significantly reducing mismatch, and drastically saving device area, complexity, and energy. The intrinsic stability of attractor points arises uniquely from the continuous-time dynamics of the DenseAM, making these models particularly well suited to analog hardware.

\subsection{Autoregressive inference} 
Dashed lines in \autoref{fig:et-circuit} illustrate the autoregressive inference procedure of the Analog ET. 
To generate the $L$-th token given context tokens $\mathbf{x}^{(1)},\dots,\mathbf{x}^{(L-1)}$, each token is first embedded and concatenated to form the attention weight matrix 
\begin{align*}
\boldsymbol{\xi}^{\text{attn},(L-1)}=\begin{bmatrix}\mathbf{e}^{(1)}\\\mathbf{e}^{(2)}\\\vdots\\\mathbf{e}^{(L-1)}\end{bmatrix}\in\mathbb R^{(L-1)\times D}
\end{align*}
These rows are loaded into the Analog ET's energy attention weight matrix $\boldsymbol\xi^\text{attn}$ by programming the corresponding crossbar resistances. During inference, the visible state $\mathbf{v}(t)$ evolves according to the Analog ET dynamics until convergence. A decoder readout (e.g., a linear layer) applied to the converged $\mathbf{v}(t=T)$ values produces logits, from which the next token $\mathbf{x}^{(L)}$ is sampled. This token is then embedded to form $\mathbf{e}^{(L)}$, and appended to the existing context. The cycle repeats with the updated attention weight matrix 
\begin{align*}
\boldsymbol{\xi}^{\text{attn},(L)}=\begin{bmatrix}\boldsymbol\xi^{\text{attn}, (L-1)}\\\mathbf{e}^{(L)}\end{bmatrix}\in\mathbb R^{L\times D}
\end{align*}
which now includes the new embedding $\mathbf{e}^{(L)}$. In hardware, this corresponds to connecting an additional hidden neuron in the energy attention block of \autoref{fig:et-circuit}, and setting its resistive weights with $\mathbf{e}^{(L)}$. Because the physical order of hidden neurons does not affect the energy function, this new neuron can be placed in any position among the hidden neurons. When the context length is fixed, the hidden neuron corresponding to the earliest token can simply be reprogrammed with the new vector of weights $\mathbf{e}^{(L)}$, resulting in the hardware equivalent of a sliding-window context. In practice, an external digital controller, e.g., an Field-Programmable Gate Array (FPGA) or  Application-Specific Integrated Circuit (ASIC) would orchestrate crossbar programming and token decoding, while the DenseAM dynamics perform the far more substantial workload of computing each next-token embedding.

This procedure is analogous to key-value (KV) caching in standard transformer inference \cite{dai2019transformer}. Context tokens $\mathbf{x}^{(1)},\dots,\mathbf{x}^{(L-1)}$ produce key and value vectors $\mathbf{k}^{(1)},\dots,\mathbf{k}^{(L-1)}$ and $\mathbf{v}^{(1)},\dots,\mathbf{v}^{(L-1)}$ respectively. When new token $\mathbf{x}^{(L)}$ is generated, its corresponding $\mathbf{k}^{(L)}$ and $\mathbf{v}^{(L)}$ vectors are appended to the cache, allowing all previous $\mathbf{k}^{(<L)}$ and $\mathbf{v}^{(<L)}$ to be reused without recomputation. When the key and value matrices are tied so that $\mathbf{k}^{(A)}=\mathbf{v}^{(A)}$, the ET's row-append operation is equivalent to the standard KV-cache update. The ET performs an autoregressive rollout that reproduces the same recurrence structure as KV-cached transformer inference, but implemented physically through the addition of new neurons and weights without touching existing hardware. For a formal derivation of the equivalence between ET attention and conventional attention with tied keys and values, see 
\cite{hoover2024energy}.

\section{Scaling properties}
Inference time and energy consumption are crucial characteristics of our system. This section investigates these metrics with respect to the network size. 

\subsection{Inference time scaling}\label{sec:inference-time-scaling}
The model (\ref{eq:xor-integrated-visible-dynamics}) and (\ref{eq: effective energy function}) is considered. In the adiabatic limit ($\tau_h\rightarrow 0$), which is satisfied by our hardware implementation, the time derivative of the energy can be written as 
\begin{equation}
    \frac{dE^\text{eff}}{dt} = \sum\limits_{i=1}^{N_v} \frac{\partial E^\text{eff}}{\partial v_i} \frac{dv_i}{dt} = - \frac{1}{\tau_v} \sum\limits_{i=1}^{N_v} \Big(\frac{\partial E^\text{eff}}{\partial v_i}\Big)^2 \sim  - \frac{N_v}{\tau_v} \label{eq: characteristic value of dE/dt}
\end{equation}
This derivative is always negative, since the dynamical system performs the gradient descent on the energy landscape. The derivative vanishes eventually when the network state vector $\boldsymbol{v}$ converges to the steady state. Since the state vector $v_i$ is typically initialized in the vicinity of the memory vectors, which are chosen to be of order one ($\sim 1$), the right hand side of (\ref{eq:xor-integrated-visible-dynamics}) is of order one too, independent of the network size. This results in the characteristic value of the temporal derivative shown in (\ref{eq: characteristic value of dE/dt}). 

At the same time, the typical value\footnote{We estimate the absolute value of the energy, since it can be both positive and negative depending on the mutual arrangement of memories, the state vector, and the number of hidden units.} of the energy (\ref{eq: effective energy function}) is 
\begin{equation}
    |E^\text{eff}| \sim N_v + \frac{1}{\beta}\log(N_h) \label{eq: characteristic value of energy E}
\end{equation}
During the inference dynamics the network is initialized in a high energy state, which has the characteristic value of energy (\ref{eq: characteristic value of energy E}), and performs energy descent to a lower value of the energy (which has a similar order of magnitude). In order to estimate the scaling of the time required to perform this energy descent, one can take a ratio of the energy drop by the rate of the energy decrease (\ref{eq: characteristic value of dE/dt}). This gives the following estimate 
\begin{equation}
    T^\text{conv} \sim \frac{|E^\text{eff}|}{\Big|\frac{dE}{dt} \Big|} \sim \tau_v\Big(1 + \frac{1}{\beta}\frac{\log(N_h)}{N_v} \Big) \sim \tau_v
\end{equation}
The last $\sim$ sign holds since in none of the designs presented here does $N_h$ grow super-exponentially in $N_v$. In fact, in all the use cases $N_h$ is sub-exponential in $N_v$. 

This back-of-the-envelope estimation provides the core intuition behind the scaling relationship. The inference time is {\bf constant}, and independent of the size of the network. A more careful analysis (\mbox{\autoref{appendix:inference-scaling}}) shows that in the high-$\beta$ regime the worst-case dependence is $\mathcal O\big(\frac{\tau_v}{\beta}\frac{\log N_h}{N_v}\big)$, which remains bounded for all architectures we consider. Thus, for our settings the convergence time is effectively constant in $N_v$ and $N_h$. Based on amplifier gain–bandwidth, slew-rate, and output-current constraints, we estimate achievable inference times of tens to hundreds of nanoseconds using existing CMOS technology (see Appendix \ref{appendix:existing-hardware}).

\subsection{Scaling of energy consumption}
We now analyze how the total inference energy scales with network size. Energy dissipation arises primarily from (i) Ohmic loss in the resistive weights, (ii) charging of neuron-state capacitors, and (iii) constant per-neuron overhead from amplifiers and bias currents. We show that, under bounded voltage swings and fixed conductance budgets, total energy grows only linearly with the number of neurons.

\paragraph{Weight dissipation.}
Let the neuron output voltages be proportional to activations: $\mathbf{u}=\kappa \mathbf{g}$ and $\mathbf{w}=\kappa \mathbf{f}$, where $\kappa$ is a fixed voltage swing. Such a bounded swing can always be enforced by global rescaling of $\boldsymbol{\xi}$, $\beta$, and voltage units without changing the dynamics (see ~\autoref{appendix:voltage-scaling}). The instantaneous power dissipated by the resistive crossbar array is
\begin{align}
P_\text{weights}(t)=\sum_{\mu=1}^{N_h}\sum_{i=1}^{N_v}\xi_{\mu i}(u_i-w_\mu)^2
\end{align}
With $0\le g_i\le1$, $\mathbf{f}$-softmax, and row/column conductance budgets $\sum_\mu\xi_{\mu i}\le C_c$, $\sum_i\xi_{\mu i}\le C_r$, the total power obeys
\begin{align}
P_\text{weights}(t)\le 2\kappa^2(C_cN_v + C_r)=\mathcal O(N_v)
\end{align}
For a runtime of duration $T\sim T^\text{conv}$, the energy dissipated by the weights is therefore $E_\text{weights}=\mathcal O(N_v T)$, where $T\sim1$ from \autoref{sec:inference-time-scaling}.

\paragraph{Capacitive and overhead energy.}
Each neuron charges a local capacitor a finite number of times by at most $V_\text{swing}\!\sim\!\kappa$, giving 
\begin{equation}
E_\text{cap} \le \kappa^2 \!\left(\sum_i C_i^{(v)} + \sum_\mu C_\mu^{(h)}\right)
= \mathcal O(N_v+N_h)
\end{equation}
Active bias and amplifier inefficiencies contribute fixed per-neuron power, yielding ${E_\text{other}=\mathcal O((N_v+N_h)T)}$.

\paragraph{Total energy scaling.}
With bounded voltage swing and conductance budgets,
\begin{equation}
    E_{\mathrm{total}} =\mathcal O(N_v + N_h)
\end{equation}
Hence, the total inference energy scales only linearly with system size. For the full derivation, see \autoref{appendix:energy-scaling}.

\subsection{Scaling of hardware area}\label{sec:scaling of area}
The area is dominated by two components: the area taken up by the synaptic weights, which is implemented as a crossbar array with programmable weights, and the area taken up by the neurons feeding the crossbar array. The area of the crossbar array scales as the number of weights $\mathcal O(N_v N_h)$. The area of the neurons scales as $\mathcal O(N_v+N_h)$.

\section{Conclusion}
In this paper, we have presented an analog accelerator architecture for Dense Associative Memories, implemented using resistive crossbar arrays and continuous-time RC neuron dynamics. Our design implements DenseAM inference as time evolution of a physical dynamical system, rather than a sequence of discrete numerical update steps. We demonstrated this architecture with three representative settings of increasing complexity: XOR, Hamming (7,4) error decoding, and an Energy Transformer-style sequence model. These examples show that the analog DenseAM accelerator architecture covers both associative memory tasks and attention-based sequence models.

Our analysis shows that DenseAM accelerators enjoy favorable asymptotic scaling properties. Inference time is constant in the dimensions of the model size, meaning that inference time is governed primarily by the physical time constants of the circuit. This is in sharp contrast to digital implementations of the same dynamics, whose runtime must grow at least linearly with model size.

To assess hardware feasibility, we derived lower bounds on the neuronal time constants imposed by amplifier gain-bandwidth product, slew rate, and output current limits in our neuron design. Reported figures from representative CMOS OTAs in the literature give inference times on the order of tens-to-hundreds of nanoseconds, even with conservative design margins. Combined with the constant scaling of inference with model size, these estimates suggest that DenseAM accelerators can match or exceed the latency of digital GPUs as models grow, without requiring exotic devices or beyond-CMOS technologies. 

Our results highlight DenseAMs as a natural abstraction for analog AI hardware.
Their error correcting dynamics and asymptotic stability directly address long-standing concerns about robustness and readout timing: small perturbations are corrected by the dynamics instead of accumulated, and the final state is stable when readout happens over a wide temporal window. At the same time, the DenseAM framework is expressive enough to capture modern primitives such as attention and transformer-like architectures, as illustrated by our Analog Energy Transformer construction. 
These properties suggest that DenseAM-based analog accelerators may be a promising substrate for future AI systems, and motivate further co-design of models, dynamics, and devices.

\subsection*{Acknowledgements}
MGB would like to thank Faiz Muhammad for exploratory attempts at SPICE simulations. DK would like to thank Kwabena Boahen for helpful discussions.

\printbibliography

@inproceedings{krotov2020large,
  title={Large Associative Memory Problem in Neurobiology and Machine Learning},
  author={Krotov, Dmitry and Hopfield, John J},
  booktitle={International Conference on Learning Representations},
  year={2021}
}

@inproceedings{hoover2022universal,
  title={A universal abstraction for hierarchical hopfield networks},
  author={Hoover, Benjamin and Chau, Duen Horng and Strobelt, Hendrik and Krotov, Dmitry},
  booktitle={The Symbiosis of Deep Learning and Differential Equations II},
  year={2022}
}

@article{aifer2025solving,
  title={Solving the compute crisis with physics-based ASICs},
  author={Aifer, Maxwell and Belateche, Zach and Bramhavar, Suraj and Camsari, Kerem Y and Coles, Patrick J and Crooks, Gavin and Durian, Douglas J and Liu, Andrea J and Marchenkova, Anastasia and Martinez, Antonio J and others},
  journal={arXiv preprint arXiv:2507.10463},
  year={2025}
}

@inproceedings{sohl2015deep,
  title={Deep unsupervised learning using nonequilibrium thermodynamics},
  author={Sohl-Dickstein, Jascha and Weiss, Eric and Maheswaranathan, Niru and Ganguli, Surya},
  booktitle={International conference on machine learning},
  pages={2256--2265},
  year={2015},
  organization={pmlr}
}

@article{krotov2016dense,
  title={Dense associative memory for pattern recognition},
  author={Krotov, Dmitry and Hopfield, John J},
  journal={Advances in neural information processing systems},
  volume={29},
  year={2016}
}

@article{ramsauer2020hopfield,
  title={Hopfield networks is all you need},
  author={Ramsauer, Hubert and Sch{\"a}fl, Bernhard and Lehner, Johannes and Seidl, Philipp and Widrich, Michael and Adler, Thomas and Gruber, Lukas and Holzleitner, Markus and Pavlovi{\'c}, Milena and Sandve, Geir Kjetil and others},
  journal={arXiv preprint arXiv:2008.02217},
  year={2020}
}

@article{hopfield1982neural,
  title={Neural networks and physical systems with emergent collective computational abilities.},
  author={Hopfield, John J},
  journal={Proceedings of the national academy of sciences},
  volume={79},
  number={8},
  pages={2554--2558},
  year={1982},
  publisher={National Acad Sciences}
}

@article{hoover2023memory,
  title={Memory in plain sight: A survey of the uncanny resemblances between diffusion models and associative memories},
  author={Hoover, Benjamin and Strobelt, Hendrik and Krotov, Dmitry and Hoffman, Judy and Kira, Zsolt and Chau, Duen Horng},
  journal={arXiv preprint arXiv:2309.16750},
  year={2023}
}

@article{ambrogioni2023search,
  title={In search of dispersed memories: Generative diffusion models are associative memory networks},
  author={Ambrogioni, Luca},
  journal={arXiv preprint arXiv:2309.17290},
  year={2023}
}

@article{vaswani2017attention,
  title={Attention is all you need},
  author={Vaswani, Ashish},
  journal={arXiv preprint arXiv:1706.03762},
  year={2017}
}

@article{krotov2021hierarchical,
  title={Hierarchical associative memory},
  author={Krotov, Dmitry},
  journal={arXiv preprint arXiv:2107.06446},
  year={2021}
}

@article{sillman2023analog,
  title={Analog Implementation of the Softmax Function},
  author={Sillman, Jacob},
  journal={arXiv preprint arXiv:2305.13649},
  year={2023}
}

@article{hopfield1986computing,
  title={Computing with neural circuits: A model},
  author={Hopfield, John J and Tank, David W},
  journal={Science},
  volume={233},
  number={4764},
  pages={625--633},
  year={1986},
  publisher={American Association for the Advancement of Science}
}

@inproceedings{graf1986vlsi,
  title={VLSI implementation of a neural network memory with several hundreds of neurons},
  author={Graf, HP and Jackel, LD and Howard, RE and Straughn, B and Denker, JS and Hubbard, W and Tennant, DM and Schwartz, D},
  booktitle={AIP conference proceedings},
  volume={151},
  number={1},
  pages={182--187},
  year={1986},
  organization={American Institute of Physics}
}

@inproceedings{jouppi2017datacenter,
  title={In-datacenter performance analysis of a tensor processing unit},
  author={Jouppi, Norman P and Young, Cliff and Patil, Nishant and Patterson, David and Agrawal, Gaurav and Bajwa, Raminder and Bates, Sarah and Bhatia, Suresh and Boden, Nan and Borchers, Al and others},
  booktitle={Proceedings of the 44th annual international symposium on computer architecture},
  pages={1--12},
  year={2017}
}

@article{assaad2009recycling,
  title={The recycling folded cascode: A general enhancement of the folded cascode amplifier},
  author={Assaad, Rida S and Silva-Martinez, Jose},
  journal={IEEE Journal of Solid-State Circuits},
  volume={44},
  number={9},
  pages={2535--2542},
  year={2009},
  publisher={IEEE}
}

@inproceedings{perez2012performance,
  title={Performance enhanced op-amp for 65nm CMOS technologies and below},
  author={Perez, Aldo Pena and Maloberti, Franco},
  booktitle={2012 IEEE International Symposium on Circuits and Systems (ISCAS)},
  pages={201--204},
  year={2012},
  organization={IEEE}
}

@article{naderi2018operational,
  title={Operational transconductance amplifier with class-B slew-rate boosting for fast high-performance switched-capacitor circuits},
  author={Naderi, Mohammad H and Prakash, Suraj and Silva-Martinez, Jose},
  journal={IEEE Transactions on Circuits and Systems I: Regular Papers},
  volume={65},
  number={11},
  pages={3769--3779},
  year={2018},
  publisher={IEEE}
}

@inproceedings{yen2020high,
  title={A High Slew Rate, Low Power, Compact Operational Amplifier Based on the Super-Class AB Recycling Folded Cascode},
  author={Yen, Alec and Blalock, Benjamin J},
  booktitle={2020 IEEE 63rd International Midwest Symposium on Circuits and Systems (MWSCAS)},
  pages={9--12},
  year={2020},
  organization={IEEE}
}

@article{krotov2018adversarial,
  title={Dense associative memory is robust to adversarial inputs},
  author={Krotov, Dmitry and Hopfield, John},
  journal={Neural computation},
  volume={30},
  number={12},
  pages={3151--3167},
  year={2018},
  publisher={MIT Press One Rogers Street, Cambridge, MA 02142-1209, USA journals-info~…}
}

@article{hoover2024energy,
  title={Energy transformer},
  author={Hoover, Benjamin and Liang, Yuchen and Pham, Bao and Panda, Rameswar and Strobelt, Hendrik and Chau, Duen Horng and Zaki, Mohammed and Krotov, Dmitry},
  journal={Advances in Neural Information Processing Systems},
  volume={36},
  year={2024}
}

@article{hamming1950error,
  title={Error detecting and error correcting codes},
  author={Hamming, Richard W},
  journal={The Bell system technical journal},
  volume={29},
  number={2},
  pages={147--160},
  year={1950},
  publisher={Nokia Bell Labs}
}

@article{masanet2020recalibrating,
  title={Recalibrating global data center energy-use estimates},
  author={Masanet, Eric and Shehabi, Arman and Lei, Nuoa and Smith, Sarah and Koomey, Jonathan},
  journal={Science},
  volume={367},
  number={6481},
  pages={984--986},
  year={2020},
  publisher={American Association for the Advancement of Science}
}

@article{patterson2021carbon,
  title={Carbon emissions and large neural network training},
  author={Patterson, David and Gonzalez, Joseph and Le, Quoc and Liang, Chen and Munguia, Lluis-Miquel and Rothchild, Daniel and So, David and Texier, Maud and Dean, Jeff},
  journal={arXiv preprint arXiv:2104.10350},
  year={2021}
}

@article{pham2025memorization,
  title={Memorization to generalization: Emergence of diffusion models from associative memory},
  author={Pham, Bao and Raya, Gabriel and Negri, Matteo and Zaki, Mohammed J and Ambrogioni, Luca and Krotov, Dmitry},
  journal={arXiv preprint arXiv:2505.21777},
  year={2025}
}

@article{krotov2025modern,
  title={Modern methods in associative memory},
  author={Krotov, Dmitry and Hoover, Benjamin and Ram, Parikshit and Pham, Bao},
  journal={arXiv preprint arXiv:2507.06211},
  year={2025}
}

@article{tang2021remark,
  title={A remark on a paper of krotov and hopfield [arxiv: 2008.06996]},
  author={Tang, Fei and Kopp, Michael},
  journal={arXiv preprint arXiv:2105.15034},
  year={2021}
}

@article{werbos2002backpropagation,
  title={Backpropagation through time: what it does and how to do it},
  author={Werbos, Paul J},
  journal={Proceedings of the IEEE},
  volume={78},
  number={10},
  pages={1550--1560},
  year={2002},
  publisher={IEEE}
}

@phdthesis{kidger2021on,
    title={{O}n {N}eural {D}ifferential {E}quations},
    author={Patrick Kidger},
    year={2021},
    school={University of Oxford},
}

@inproceedings{dai2019transformer,
  title={Transformer-xl: Attentive language models beyond a fixed-length context},
  author={Dai, Zihang and Yang, Zhilin and Yang, Yiming and Carbonell, Jaime G and Le, Quoc and Salakhutdinov, Ruslan},
  booktitle={Proceedings of the 57th annual meeting of the association for computational linguistics},
  pages={2978--2988},
  year={2019}
}

@book{mead2012analog,
  title={Analog VLSI implementation of neural systems},
  author={Mead, Carver and Ismail, Mohammed},
  volume={80},
  year={2012},
  publisher={Springer Science \& Business Media}
}

@article{hopfield1990effectiveness,
  title={The effectiveness of analogue'neural network'hardware},
  author={Hopfield, JJ},
  journal={Network: Computation in Neural Systems},
  volume={1},
  number={1},
  pages={27},
  year={1990},
  publisher={IOP Publishing}
}

@article{musa2025dense,
  title={Dense Associative Memory in a Nonlinear Optical Hopfield Neural Network},
  author={Musa, Khalid and Kumar, Santosh and Katidis, Michael and Huang, Yu-Ping},
  journal={arXiv preprint arXiv:2506.07849},
  year={2025}
}

@article{marsh2021enhancing,
  title={Enhancing associative memory recall and storage capacity using confocal cavity QED},
  author={Marsh, Brendan P and Guo, Yudan and Kroeze, Ronen M and Gopalakrishnan, Sarang and Ganguli, Surya and Keeling, Jonathan and Lev, Benjamin L},
  journal={Physical Review X},
  volume={11},
  number={2},
  pages={021048},
  year={2021},
  publisher={APS}
}

@incollection{tank1988simple,
  title={Simple “Neural” optimization networks: an A/D converter, signal decision circuit, and a linear programming circuit},
  author={Tank, David W and Hopfield, John J},
  booktitle={Artificial neural networks: theoretical concepts},
  pages={87--95},
  year={1988}
}

@article{hopfield1984neurons,
  title={Neurons with graded response have collective computational properties like those of two-state neurons.},
  author={Hopfield, John J},
  journal={Proceedings of the national academy of sciences},
  volume={81},
  number={10},
  pages={3088--3092},
  year={1984}
}

@article{guo2015modeling,
  title={Modeling and experimental demonstration of a Hopfield network analog-to-digital converter with hybrid CMOS/memristor circuits},
  author={Guo, Xinjie and Merrikh-Bayat, Farnood and Gao, Ligang and Hoskins, Brian D and Alibart, Fabien and Linares-Barranco, Bernabe and Theogarajan, Luke and Teuscher, Christof and Strukov, Dmitri B},
  journal={Frontiers in neuroscience},
  volume={9},
  pages={488},
  year={2015},
  publisher={Frontiers Media SA}
}

@article{eryilmaz2014brain,
  title={Brain-like associative learning using a nanoscale non-volatile phase change synaptic device array},
  author={Eryilmaz, Sukru B and Kuzum, Duygu and Jeyasingh, Rakesh and Kim, SangBum and BrightSky, Matthew and Lam, Chung and Wong, H-S Philip},
  journal={Frontiers in neuroscience},
  volume={8},
  pages={205},
  year={2014},
  publisher={Frontiers Media SA}
}

@article{hu2015associative,
  title={Associative memory realized by a reconfigurable memristive Hopfield neural network},
  author={Hu, SG and Liu, Y and Liu, Z and Chen, TP and Wang, JJ and Yu, Q and Deng, LJ and Yin, Y and Hosaka, Sumio},
  journal={Nature communications},
  volume={6},
  number={1},
  pages={7522},
  year={2015},
  publisher={Nature Publishing Group UK London}
}

@article{schlogl2007design,
  title={A design example of a 65 nm CMOS operational amplifier},
  author={Schl{\"o}gl, Franz and Zimmermann, Horst},
  journal={International Journal of Circuit Theory and Applications},
  volume={35},
  number={3},
  pages={343--354},
  year={2007},
  publisher={Wiley Online Library}
}

\appendix
\section{Neuron Design}\label{appendix:neuron-design}
\begin{figure}
    \centering
    \includegraphics[width=0.75\linewidth]{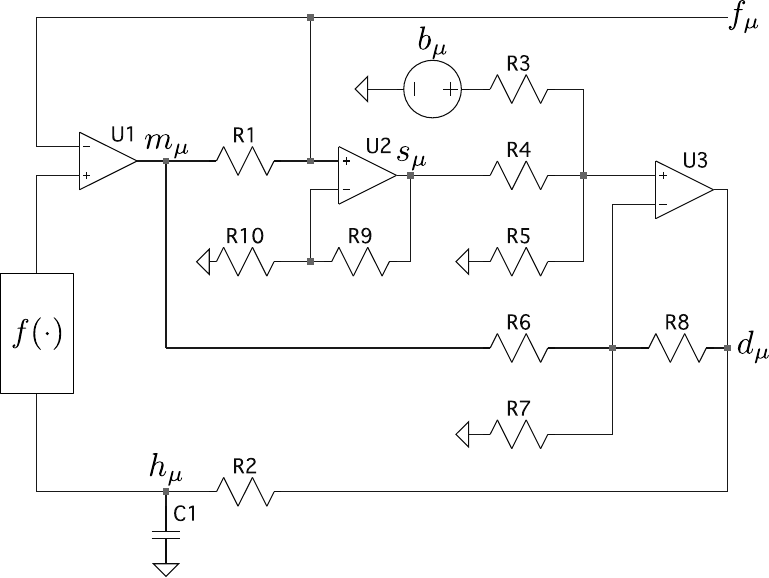}
    \caption{Circuit for a single neuron.}
    \label{fig:single-neuron-circuit}
\end{figure}
Figure~\ref{fig:single-neuron-circuit} shows the circuit design of a single neuron, with labels corresponding to this being a hidden neuron at index $\mu$. We derive the dynamics of the neuron internal state $h_\mu$ and activation output voltage $f_\mu$. We proceed using only Kirchhoff's Current Law (KCL) and the definition of an ideal op-amp.

\paragraph{Assumptions and conventions.}
\begin{itemize}
\item \textbf{Ideal op-amps:} infinite open-loop gain, infinite input impedance (no input current), zero output impedance. Under stable negative feedback this enforces a virtual short $V_+=V_-$.
\item \textbf{Current $J_\mu$:} we define $J_\mu$ as the current which flows from $f_\mu$ to $m_\mu$ through $R_1$.
\item \textbf{Op-amp input labels:} We denote the inverting and noninverting inputs of each op-amp explicitly, e.g.\ $U2_-$ for the inverting input of \texttt{U2}, $U3_+$ for the noninverting input of \texttt{U3}, etc. 
\item \textbf{Node labels:} Label $m_\mu$ as the output of \texttt{U1}, $s_\mu$ as the output of \texttt{U2}, and $d_\mu$ as the output of \texttt{U3}. The neuron pre-activation state is labeled $h_\mu$, and the post-activation state is labeled $f_\mu$. Voltage $b_\mu$ (as an ideal voltage source) drives the bias for this neuron. Voltages $h_\mu$, $b_\mu$, and $f_\mu$ correspond directly to the state variables in equation~\eqref{eq:original differential equations}. 
\end{itemize}

\paragraph{Block \texttt{U1}: buffer of activation voltage $f_\mu$.}
Op-amp \texttt{U1} buffers the output of the activation function $f(\cdot)$ and drives the output of the neuron, $f_\mu$. Because no current can flow into $U1_-$, all the current flowing into this neuron must flow through $R_1$ to $m_\mu$ and is sourced or sunk by \texttt{U1}'s output node.

\paragraph{Block \texttt{U2}: non-inverting stage producing $s_\mu$ from $f_\mu$ and $m_\mu$.}
The positive input of \texttt{U2} is $U2_{+}=f_\mu$, and by \texttt{U2}'s virtual short, the negative input $U2_-=U2_+=f_\mu$. By KCL at $U2_-$,
\begin{align}
    \frac{U2_-}{R_{10}}=\frac{s_\mu-U2_-}{R_9} \quad \Rightarrow \quad s_\mu=\left(1+\frac{R_9}{R_{10}}\right)f_\mu
    \label{eq:kcl_s_mu}
\end{align}

\paragraph{Block \texttt{U3}: non-inverting stage producing $d_\mu$ from $s_\mu$, $b_\mu$, and $m_\mu$.}
By KCL at the positive input of \texttt{U3},
\begin{align}
    \frac{b_\mu-U3_+}{R_3}+\frac{s_\mu-U3_+}{R_4}=\frac{U3_+}{R_5}\quad \Rightarrow\quad U3_+= \frac{R_4 R_5 b_\mu + R_3 R_5 s_\mu}{R_4 R_5 + R_3 R_5 + R_3 R_4}
    \label{eq:kcl_U3_+}
\end{align}
KCL at the negative input of \texttt{U3} gives us
\begin{align}
    \frac{m_\mu-U3_-}{R_6}+\frac{-U3_-}{R_7}=\frac{U3_--d_\mu}{R_8}\quad\Rightarrow\quad d_\mu=U3_-\left(1+R_8\left(\frac1{R_6}+\frac1{R_7}\right)\right)-\frac{R_8 m_\mu}{R_6}
    \label{eq:kcl_d_mu}
\end{align}
Virtual short of \texttt{U3} means $U3_-=U3_+$. Combining equations \eqref{eq:kcl_U3_+} and \eqref{eq:kcl_d_mu}, get
\begin{align}
    d_\mu &= \frac{R_6 R_7 + R_8 (R_6 + R_7)}{R_6 R_7} \cdot \frac{R_4 R_5 b_\mu + R_3 R_5 s_\mu}{R_4 R_5 + R_3 R_5 + R_3 R_4} - \frac{R_8}{R_6} m_\mu
\label{eq:d_mu_full}
\end{align}

\paragraph{Dynamics of RC circuit.}
$R_2$ and $C_1$ form an RC circuit driven by voltage $d_\mu$. The voltage across the capacitor $h_\mu$ follows the relation
\begin{align}
    R_2C_1\frac{dh_\mu}{dt}&=-h_\mu+d_\mu \nonumber\\
    &=-h_\mu+\frac{R_6 R_7 + R_8 (R_6 + R_7)}{R_6 R_7} \cdot \frac{R_4 R_5 b_\mu + R_3 R_5 s_\mu}{R_4 R_5 + R_3 R_5 + R_3 R_4} - \frac{R_8}{R_6} m_\mu
    \label{eq:rc_circuit_full}
\end{align}

\paragraph{With incoming current.}
Take the incoming current $J_\mu=\sum_i \xi_{\mu i}(g_i-f_\mu)$. This produces a voltage drop across $R_1$ such that $m_\mu=f_\mu-R_1J_\mu=f_\mu-R_1\sum_i \xi_{\mu i}(g_i-f_\mu)$. Then, the dynamics of $h_\mu$ from equation~\eqref{eq:rc_circuit_full} are
\begin{align}
    R_2C_1\frac{dh_\mu}{dt}
    &= -h_\mu+\frac{R_6 R_7 + R_8 (R_6 + R_7)}{R_6 R_7} \cdot \frac{R_4 R_5 b_\mu + R_3 R_5 s_\mu}{R_4 R_5 + R_3 R_5 + R_3 R_4} - \frac{R_8}{R_6} \left(f_\mu-R_1J_\mu\right)
    \label{eq:dh_mu_dt_j}
\end{align}
Substituting in $s_\mu$ from equation~\eqref{eq:kcl_s_mu} and $J_\mu$:
\begin{align}
R_2C_1\frac{dh_\mu}{dt}
    &=-h_\mu+\frac{R_6 R_7 + R_8 (R_6 + R_7)}{R_6 R_7} \cdot \frac{R_4 R_5 b_\mu + R_3 R_5 \left(1+\frac{R_9}{R_{10}}\right)f_\mu}{R_4 R_5 + R_3 R_5 + R_3 R_4} - \frac{R_8}{R_6} \left(f_\mu-R_1\sum_i \xi_{\mu i}(g_i-f_\mu)\right)
    \label{eq:dh_mu_dt_full}
\end{align}

\paragraph{Equal-resistance special case.}
Set $R_1=R_3=R_4=R_5=R_6=R_7=R_8$. Then, equation~\eqref{eq:dh_mu_dt_full} reduces to
\begin{align}
    R_2C_1\frac{dh_\mu}{dt}=-h_\mu+b_\mu+\frac{R_9}{R_{10}}f_\mu+\sum_i \xi_{\mu i}(g_i-f_\mu)
    \label{eq:dh_mu_dt_equal_res}
\end{align}

\paragraph{Selection of $R_9/R_{10}$ self-term gain.}
Evidently, in order to match the form of equation~\eqref{eq:original differential equations}, we need to cancel the $-f_\mu\sum_i \xi_{\mu i}$ term that appears on the right hand side of equation~\eqref{eq:dh_mu_dt_equal_res}. The $R_9/R_{10}$ term allows us to do that by setting
\begin{align}
    \frac{R_9}{R_{10}}=\sum_i \xi_{\mu i}
    \label{eq:r7_r8_assignment}
\end{align}
Taking equation~\eqref{eq:r7_r8_assignment}'s assignment to $R_9$ and $R_{10}$ simplifies equation~\eqref{eq:dh_mu_dt_equal_res} into
\begin{align}
    R_2C_1\frac{dh_\mu}{dt}&=\sum_i \xi_{\mu i}g_i -h_\mu + b_\mu
    \label{eq:dh_mu_dt}
\end{align}
which exactly matches our desired dynamics.

\subsection{Activation function}
The voltage across $C_1$ gives us the dynamics of the neuron internal state $h_\mu$. Figure~\ref{fig:single-neuron-circuit} contains a block representing a nonlinear amplifier, denoted $f(\cdot)$, whose input is $h_\mu$ and whose output is $f_\mu=f(h_\mu)$. This voltage is buffered with \texttt{U1} onto the neuron output line, labeled $f_\mu$, which is what other neurons ``see'' in the crossbar array. The chosen activation function does not affect the rest of the dynamics of the neuron. Particularly, the activation function need not be element-wise: a vector-wise activation function like softmax can be readily applied instead.

\subsection{Neurons interacting in a network}
\begin{figure}
    \centering
    \includegraphics[width=0.5\linewidth]{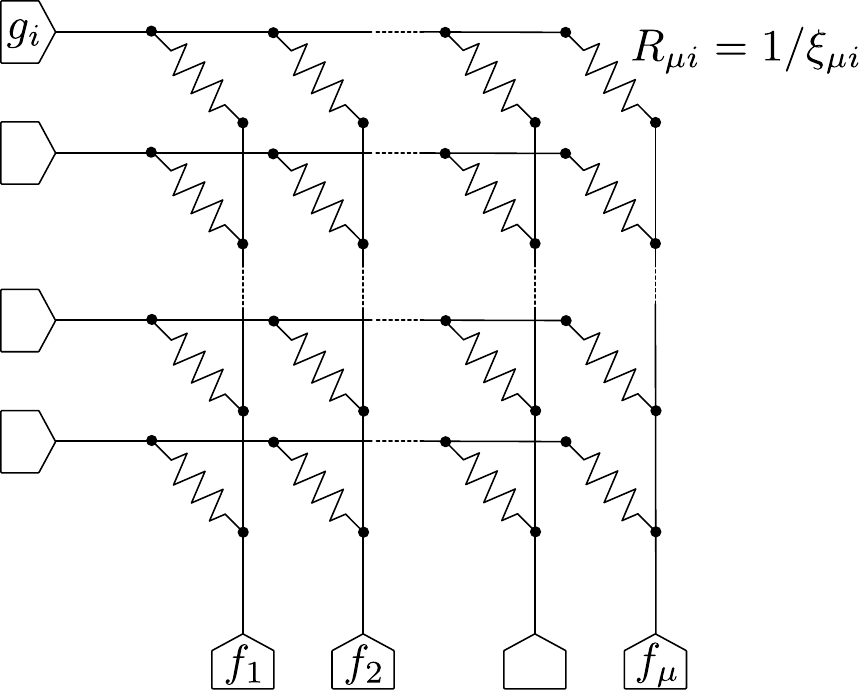}
    \caption{Crossbar Array. Each pentagon contains a neuron of design in Figure~\ref{fig:single-neuron-circuit}. In this layout we have flipped the crossbar array, so that index $\mu$ runs horizontally and index $i$ runs vertically.}
    \label{fig:crossbar}
\end{figure}
So far we have examined the dynamics of a single neuron, treating as an assumption that the neuron will receive an incoming current $J_\mu=\sum_i \xi_{\mu i}(g_i-f_\mu)$. Now, we will show how to wire these neurons together to realize this. Figure~\ref{fig:crossbar} shows the simplest DenseAM construction where each pentagonal node is a circuit of design in Figure~\ref{fig:single-neuron-circuit}. Each neuron exposes a single node whose voltage is driven at the activation of the neuron, and which accepts an incoming current which it uses to drive its dynamics. Each hidden neuron $f_\mu$ is connected to a visible neuron $g_i$ via a resistance $R_{\mu i}=1/\xi_{\mu i}$ that is the inverse of the weight it represents. The current flowing into node $f_\mu$ is  $J_\mu=\sum_i\frac{1}{R_{\mu i}}(g_i-f_\mu)$, which is the assumption needed for equation~\eqref{eq:dh_mu_dt_j}. This same analysis holds for other hidden and visible neurons, and so together they realize the large dynamical system of~\eqref{eq:original differential equations}.

\subsection{SPICE Netlist}
Following is the SPICE netlist for the single neuron circuit, using ideal op-amps. Component values are omitted for brevity. There is no nonlinearity here; adding one would be a matter of inserting a nonlinear amplifier between node \texttt{h\_µ} and \texttt{XU1}'s positive terminal. 
\begin{verbatim}
R1 f_µ m_µ
XU1 f_µ h_µ m_µ opamp Aol=100K GBW=10Meg
XU2 u2- f_µ s_µ opamp Aol=100K GBW=10Meg
R2 u2- 0
R3 s_µ u2-
R4 u3+ s_µ
R5 u3+ 0
XU3 u3- u3+ d_µ opamp Aol=100K GBW=10Meg
R6 u3- m_µ
R7 d_µ u3-
R8 d_µ h_µ
C1 h_µ 0
V§b_µ N001 0
R9 u3+ N001
R10 u3- 0
\end{verbatim}

\section{Softmax Circuit}\label{appendix:softmax-circuit}
\begin{figure}
    \centering
    \includegraphics[width=0.75\linewidth]{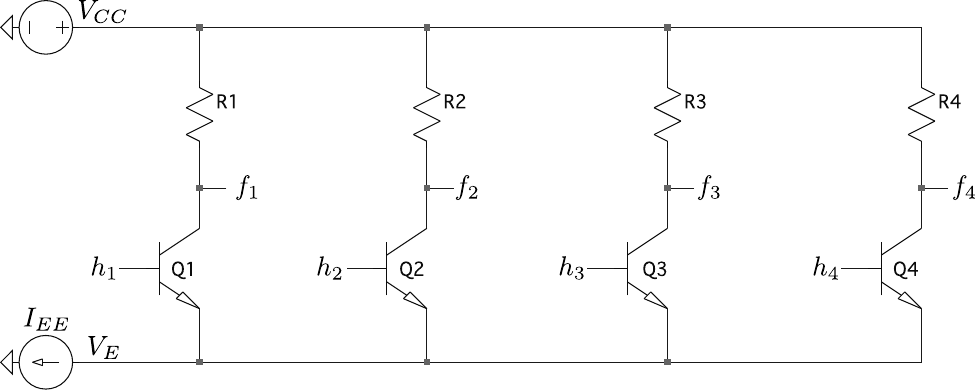}
    \caption{Softmax circuit design}
    \label{fig:softmax}
\end{figure}

For demonstration purposes, we follow the construction of an analog softmax circuit using bipolar junction transistors (BJTs) described in~\cite{sillman2023analog}. Figure~\ref{fig:softmax} shows the design of a four-way softmax circuit using BJTs. The softmax function we aim to produce is:
\begin{align}
    \mathrm{softmax}_i = \frac{e^{z_i}}{\sum_{j=1}^{N} e^{z_j}}, \qquad i = 1, \dots, N
\end{align}

For the $\mu$th BJT in the circuit, the collector current $I_{C,\mu}$ can be expressed in terms of the base voltage $h_\mu$ and the emitter voltage $V_E$ when in the forward-active mode as:
\begin{align}
    I_{C,\mu}=I_s e^{V_{BE}/V_T},\quad V_{BE,\mu}=h_\mu-V_E,\quad \Rightarrow\quad I_{C,\mu}=I_S e^{\frac{h_\mu-V_E}{V_T}}\label{eq:collector-current}
\end{align}
where $I_s$ is the BJT's saturation current and $V_T$ is the thermal voltage. Assuming large BJT $\beta$ (note: this $\beta$ is unrelated to the softmax $\beta$)\footnote{In BJTs, $\beta$ denotes the ratio of the collector current to the base current. High BJT $\beta$ indicates the transistor is able to amplify a small base current into a much larger collector current, allowing the BJT to function as an amplifier or switch. A high $\beta$ reflects that the BJT can efficiently transmit carriers from emitter to collector, without losing them to the base.}, we can neglect base currents $I_{C,\mu}=I_{E,\mu}$. Applying KCL at the shared emitter node $V_E$, the total current $I_{EE}=\sum_{\mu=1}^N I_{C,\mu}$. We can expand the expression for the collector currents to get the currents in terms of node voltages:
\begin{align}
    I_{EE}
    &= \sum_{\mu=1}^{N_h} I_S e^{(h_\mu-V_E)/V_T}\nonumber\\
    &= \sum_{\mu=1}^{N_h} \frac{I_S e^{h_\mu/V_T}}{e^{V_E/V_T}}\label{eq:drain-current}
\end{align}
Simultaneously, the current $I_{EE}$ is also fixed by the ideal current source, so $I_{C,\mu}$ can also be expressed as the ratio of the branch current to the total current: $I_{C,\mu}=\frac{I_{C,\mu}}{I_{EE}}I_{EE}$. Plugging in \eqref{eq:collector-current} for $I_{C,\mu}$ and \eqref{eq:drain-current} for $I_{EE}$ in the denominator and canceling the term containing $V_E$,
\begin{align}
    I_{C,\mu}&=\frac{e^{h_\mu/V_T}}{\sum_{j=1}^{N_h} e^{h_j/V_T}}I_{EE}
\end{align}
This already looks very much like the ideal softmax function. The voltage at node $f_i$ is created by current flowing through resistor $R_i$, producing a voltage drop relative to $V_{CC}$. Specifically, the voltage $f_\mu=V_{CC}-\frac{e^{h_\mu/V_T}}{\sum_{j=1}^{N_h} e^{h_j/V_T}}I_{EE}R_\mu$. When $I_{EE} R_\mu=1$, this voltage $f_\mu$ is a negated and shifted softmax in the range of 1 volt. This scale and negation can be easily corrected with an op amp, which is also needed to isolate the node and prevent loading. Note that $V_{CC}$ must be chosen to be positive supply in order for the BJTs to remain in the forward-active mode. 

\section{XOR DenseAM Circuit}\label{sec:xor-spice}
\begin{sidewaysfigure}
  \centering
  \includegraphics[width=\textheight]{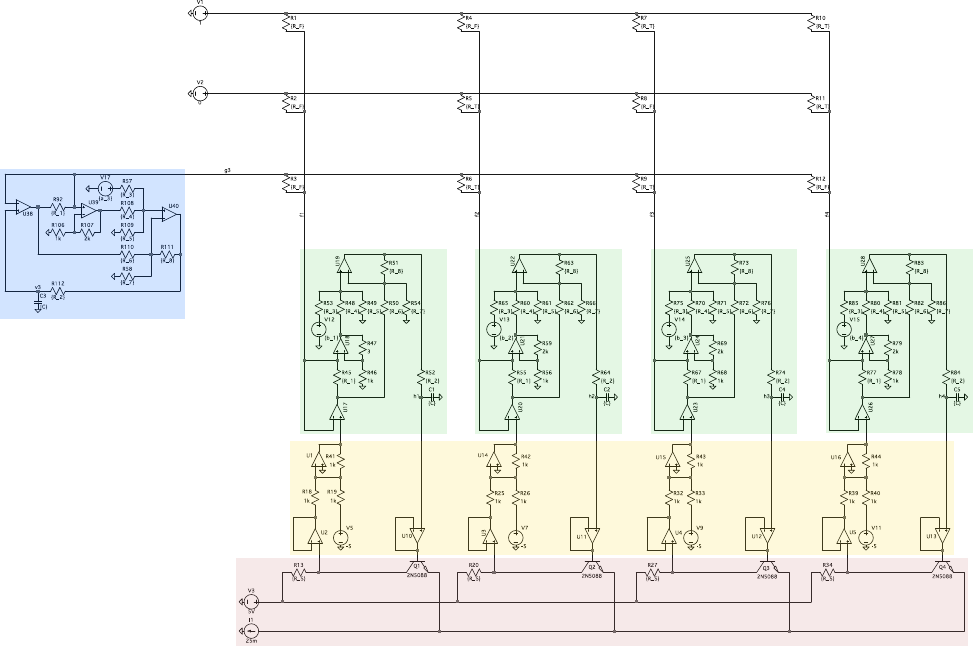}
  \caption{Full schematic for XOR DenseAM built with 1 evolving linear visible neuron and 4 hidden neurons with softmax activation. \textbf{Blue:} visible neuron. \textbf{Green:} hidden neurons. \textbf{Yellow:} buffers for softmax activation circuit. \textbf{Red:} analog softmax circuit.}
  \label{fig:xor-dam-circuit}
\end{sidewaysfigure}

Figure~\ref{fig:xor-dam-circuit} is a full circuit diagram of the DenseAM that solves the XOR problem. Given input voltages at V1, V2$ \in\{0, 1\}$, the output voltage at g3 is the result of the XOR operation between V1 and V2. In this model, the visible neuron is linear, and the hidden neurons share a softmax activation function implemented by a set of bipolar junction transistors.
Table \ref{tab:xor-parameter-values} lists the component values used in simulation. 
\begin{table}
\centering
\begin{tabular}{l
                S[table-format=3.0]
                l}
\toprule
\textbf{Parameter} & \textbf{Value} & \\
\midrule
$R_F$ & 1000 & \si{\ohm} \\
$R_T$ & 1    & \si{\ohm} \\
$R_1$ & 1    & \si{\ohm} \\
$R_2,R_3,\dots,R_8$ & 10000 & \si{\ohm} \\
$R_S$ & 40   & \si{\ohm} \\
$C$   & 10   & \si{\micro\farad} \\
$a_3$ & 0    & \si{\volt} \\
$b_1$ & 0    & \si{\volt} \\
$b_2$ & -1   & \si{\volt} \\
$b_3$ & -1   & \si{\volt} \\
$b_4$ & -1   & \si{\volt} \\
\bottomrule
\end{tabular}
\caption{Component and parameter values.}
\label{tab:xor-parameter-values}
\end{table}

\paragraph{Visible neurons.}
In the XOR task, only one visible neuron is left evolving, corresponding to the output column of the truth table. As such, the first two neurons are clamped to the input voltages, represented by \texttt{V1} and \texttt{V2}. The third visible neuron, highlighted in blue, is a linear unit with no nonlinear activation: the internal state voltage $v_3$ directly drives the output, setting $g_3=v_3$. This is the same circuit described in Appendix~\ref{appendix:neuron-design}, except where the activation block is not present.

\paragraph{Hidden neurons.} 
The XOR task requires four hidden neurons, highlighted in green. These are identical circuit constructions with the exception of the voltage sources $b_\mu$ for the biases, which are set according to the values in Table~\ref{tab:xor-parameter-values}. Unlike the visible neuron, the hidden neurons have a softmax activation function, such that $f_\mu=\text{softmax}_\mu(h)$. 

\paragraph{Softmax activation function.}
The red highlights the same softmax circuit described in Appendix~\ref{appendix:softmax-circuit}, comprised of BJT transistors, resistors, a voltage source for $V_{CC}$ and a current source for $I_{EE}$. We use the \texttt{2N5088} transistors in our model, reflecting a standard and widely available BJT. Noninverting buffers (\texttt{U10}, \texttt{U11}, etc.) are used to prevent loading effects on the state capacitors $C_\mu$ from current draw of the BJT base in forward-active mode. As discussed in Appendix~\ref{appendix:softmax-circuit}, the softmax circuit itself produces an output voltage of
\begin{align*}
    \mathrm{softmax}(z)_i = V_{CC}-\frac{e^{z_i}}{\sum_{j=1}^{N} e^{z_j}}, \qquad i = 1, \dots, N
\end{align*}
When $V_{CC}=5V$ as in this circuit, this requires extra circuitry, highlighted in yellow, to shift and negate the softmax output. This is done by first buffering the voltage output to prevent loading effects, followed by a summing op amp that subtracts $V_{CC}$ and inverts the softmax output. For the first hidden neuron $h_1$ (lower left of figure), op-amp \texttt{U2} buffers the voltage output, while \texttt{U1} is configured in an inverting summing configuration to add -5V (the inverse of $V_{CC}$) to the buffered voltage output, producing the correct softmax output. 

\paragraph{Weight matrix.}
The weight matrix is comprised of resistors $R_1$-$R_{12}$ that represent the weight matrix $\boldsymbol\xi$. These are set directly according to the XOR truth table, where each row corresponds to one hidden neuron. A boolean value of 1 ($R_T$) is set to be a high conductance ($1\Omega$), while a boolean value of 0 ($R_F$) is set to be a relatively small conductance ($1\rm k\Omega$).

The gain $s_i/g_i$ governing the value of $s_i$ is set to be the sum of the resistances in that neuron's crossbar column. The column of resistances for neuron 1 has 3 $R_F$ resistances, which sum to $3\times10^{-3}$. Hence, neuron 1's $R_{47}/R_{R46}=3/1000$. The crossbar resistances for neuron 2, 3, and 4 have 2 $R_T$ resistances and one $R_F$ resistance, which sums to approximately $2$. Hence, we approximate $R_{59}/R_{56}=2000/1000$ and similarly for hidden neurons 3 and 4.

\section{Design and implementation variations}
A large design space remains open across analog electronics and other substrates for realizing DenseAMs, with clear speed–energy–area–precision trade-offs. In electronics, the core primitives admit multiple realizations: passive, nonvolatile weights (e.g., memristors, triode-region or floating-gate transistors, and other programmable conductors); active, gained weights via OTAs; and nonlinearities via diode clamps, reverse-biased diode/BJT exponentials, MOS quadratic regions, or translinear blocks. Architectures in the spirit of \cite{hopfield1986computing, graf1986vlsi} are compact but couple synaptic values to neuronal time constants, making dynamics drift when a single weight changes—problematic for learning and consistent timing—whereas our decoupled neuron preserves a fixed time constant under weight updates. Simpler neuron/network topologies likely exist and can be attractive in resource-constrained regimes, provided their deviations from the target ODEs are validated not to degrade performance. Beyond CMOS, photonics (e.g., overdamped, low-Q microring resonators) can naturally implement first-order ODEs and can offer extreme bandwidth with distinct calibration and noise constraints. Across these options, open problems include robust weight storage/programmability and drift control, mixed-signal learning rules compatible with device limits, scaling under current/GBW/SR constraints, tolerance to mismatch/noise, and algorithm–circuit co-design to exploit substrate-specific advantages.

\section{Scaling of inference time} \label{appendix:inference-scaling}
There are two conditions under which inference times should be studied, dependent on the softmax temperature $\beta$. In the low-$\beta$ regime, the DenseAM reaches equilibria with multiple hidden neurons ``competing" in the softmax, while in the high-$\beta$ regime, the DenseAM reaches equilibria with only one hidden neuron ``winning out" in the softmax. Intuitively, the high-$\beta$ regime corresponds to exact memory recall, while the low-$\beta$ regime corresponds to interpolation. The XOR and Hamming (7,4) code are in the high-$\beta$ regime, while the energy transformer lies in the low-$\beta$ regime. In both regimes, we find that the DenseAM converges in time that is constant with respect to the number of neurons.

\paragraph{Assumptions.}

\begin{description}
  \item[(A1)] There is a per-synapse device limit of $0\le\xi_{\mu i}\le G_{\text{max}}$ where $G_\text{max}$ is the maximum conductance set by the physics of the crossbar crosspoints. Because $f$ is the output of a softmax so $f_\mu\le1~\forall \mu$, this means
    \begin{align}
        \sum_\mu\xi_{\mu i}f_\mu\le G_\text{max}
    \end{align}
    so the RHS of the visible neuron dynamics is $\mathcal O(1)$.

    There exist both column-sum and row-sum budgets that are enforced by the hardware, since each neuron's output stage can only source/sink a finite amount of current while maintaining GBW/SR margins. This dictates a per-column and per-row conductance budget to stay within this maximum current, resulting in
    \begin{align}
        \sum_i^{N_v}\xi_{\mu i}\le C_r \quad\forall\mu,\qquad \sum_\mu^{N_h}\xi_{\mu i}\le C_c\quad \forall i
    \end{align}
    
    Weights can only be positive since conductances can only be positive, so $\xi_{\mu i}\ge0$. 

    As a corollary of (A1), note also that we can bound $\|\boldsymbol\xi_\mu\|_2\le S~~\forall\mu$, and since ${\|\boldsymbol\xi_\mu\|_2\le \|\boldsymbol\xi_\mu\|_1}$, then $\|\boldsymbol\xi_\mu\|_2\le C_c~~\forall \mu$.
  \item[(A2)] Bounded biases. $|a_i|\le A$, $|b_\mu|\le B$ for all $i,\mu$. In realistic regimes, this typically holds, for example the typical choice in boolean functions of $b_\mu=-\frac\beta2\|\boldsymbol\xi_\mu\|^2$ (seen in Section~\ref{sec:xor}).
\end{description}

\paragraph{Model.}
Take the system of equation~\eqref{eq:original differential equations} with a softmax activation on hidden neurons and an identity activation on visible neurons. For clarity we assume 0 biases on visible neurons, but they do not change the analysis.
\begin{align}
    \tau_v \dot{\mathbf v} = \boldsymbol \xi^\top \mathbf f+\mathbf a-\mathbf v,\quad
    \tau_h \dot{\mathbf h} = \boldsymbol \xi \mathbf v +\mathbf b - \mathbf h ,\quad
   \mathbf  f=\text{softmax}_\beta (\mathbf h)
    \label{eq:model-dynamics}
\end{align}
Integrating out the hidden units,
\begin{align}
    \tau_v \dot{\mathbf v}
        &= \boldsymbol\xi^\top f(\mathbf v) - \mathbf v, \\
    f(\mathbf v)
        &= \text{softmax}\bigl(\beta(\boldsymbol\xi \mathbf v + \mathbf b)\bigr)
\end{align}
yields the effective energy function expressed in terms of visible neurons:
\begin{align}
    E(\mathbf v)=\frac12\|\mathbf v\|^2-\frac1\beta\log\sum_\mu\exp\left(\beta\left(\boldsymbol\xi_\mu^\top \mathbf v+\mathbf b    \right)\right)
    \label{eq:energy-in-visible-neurons}
\end{align}
where $\nabla E(\mathbf v)=\mathbf v-\boldsymbol\xi^\top f(\mathbf v)$. Because $\tau_v\dot {\mathbf v}=-\nabla E(\mathbf v)$, we see that the dynamical trajectory causes the energy to monotonically decrease over time:
\begin{align}
    \frac{d}{dt}E(\mathbf v(t))=\nabla E(\mathbf v(t))^\top\dot{\mathbf v}=-\frac1{\tau_v}\|\nabla E(\mathbf v(t))\|^2\le0
    \label{eq:energy-decay}
\end{align}

\subsection{Low-$\beta$ regime}\label{sec:low-beta}
The energy landscape in the low-$\beta$ regime exhibits uniform strong convexity, so the gradient flow dynamics cause the energy gap to decay exponentially, reaching an $\epsilon$-fraction of the original energy gap in constant time.
To show $E(\mathbf v)$ is $\alpha$-strongly convex, we must show $\nabla^2 E(\mathbf v)\succeq \alpha I$ for some $\alpha>0$. This means that all the eigenvalues of the Hessian are $\ge\alpha$. Equivalently, $\lambda_\text{min}(\nabla^2E)\ge\alpha$. Denote $G(\mathbf f)=\text{Diag}(\mathbf f) - \mathbf f\mathbf f^\top \succeq 0$, which is the Jacobian of the softmax function ${f(\mathbf v)=\text{softmax}(\beta(\boldsymbol\xi \mathbf v+\mathbf b))}$. 
\begin{align}
    \nabla^2 E(\mathbf v) &= I - \beta\boldsymbol\xi^\top G(\mathbf f)\boldsymbol\xi\\
    \lambda_{\min}\left(\nabla^2 E(\mathbf v)\right)
        &= \lambda_{\min}\left(I - \beta\boldsymbol\xi^\top G(\mathbf f)\boldsymbol\xi\right) \\
        &= 1 - \beta\lambda_{\max}\left(\boldsymbol\xi^\top G(\mathbf f)\boldsymbol\xi\right) \\
    \Rightarrow\quad 
    \nabla^2 E(\mathbf v) &\succeq \left(1 - \beta\lambda_{\max}\left(\boldsymbol\xi^\top G(\mathbf f)\boldsymbol\xi\right)\right) I
\end{align}
Because $G(\mathbf f)\preceq\text{Diag}(\mathbf f)\preceq I$ is PSD and therefore $\boldsymbol\xi G(\mathbf f)\boldsymbol\xi^\top$ is also PSD, and $G(\mathbf f)$ is a probability-weighted covariance where $\sum_\mu f_\mu=1$,
\begin{align}
    \lambda_\text{max}(\boldsymbol\xi^\top G(\mathbf f)\boldsymbol\xi)\le\text{tr}(\boldsymbol\xi^\top G(\mathbf f)\boldsymbol\xi)\le\sum_\mu f_\mu\|\boldsymbol\xi_\mu\|^2\le\max_\mu \|\boldsymbol\xi_\mu\|^2
\end{align}
Denote $S^2=\max_\mu\|\boldsymbol\xi_\mu\|^2\le C_c$ as in \textbf{(A1)}. Therefore, the Hessian of $E$ can be bounded as
\begin{align}
    \nabla^2 E(\mathbf v)\succeq (1-\beta S^2)I= \alpha I
\end{align}
where $\alpha=1-\beta S^2$. Then $\alpha>0$ when $\beta < 1/\max_\mu\|\boldsymbol\xi_\mu\|^2$. This is a sufficient (but not necessary) condition for the system to be in the low-$\beta$ (uniformly convex) regime, where the softmax is diffuse enough that its covariance term does not contribute so much negative curvature as to overwhelm the positive curvature contributed by the identity term. In this regime, the uniform lower bound on the Hessian implies $\alpha$-strong convexity, which gives the PL inequality
\begin{align}
    \frac12\|\nabla E(\mathbf v)\|^2\ge\alpha(E(\mathbf v)-E^*)
\end{align}
Together with \eqref{eq:energy-decay}, this allows us to bound the time constant of gradient flow:
\begin{align}
    \frac{d}{dt}(E(\mathbf v(t))-E^\star)=-\frac1{\tau_v}\|\nabla E(\mathbf v(t))\|^2\le -\frac{2\alpha}{\tau_v}(E(\mathbf v(t))-E^\star)
\end{align}
If the curvature is bounded below by $\alpha$, then the gradient magnitude grows at least linearly with distance to the minimum, ensuring the energy function is ``steep enough" to ensure exponential convergence. Integrating,
\begin{align}
    E(\mathbf v(t))-E^\star\le (E(\mathbf v(0))-E^\star)e^{-\frac{2\alpha}{\tau_v}t}
\end{align}
This indicates exponential decay of the energy gap. In order to reach an $\epsilon$-fraction of the original energy gap, this takes time
\begin{align}
    T(\epsilon)\le\frac{\tau_v}{2\alpha}\log\frac1\epsilon =\mathcal O(\tau_v\log(1/\epsilon))
\end{align}
which is entirely independent of system size $N_v$ and $N_h$. In the energy transformer case, this means that convergence time is entirely independent of context length $L$ and token dimension $D$. 

\subsection{High-$\beta$ regime}
\subsubsection{$T_I$: Basin selection}
Denote
\begin{align}
    s_\mu(\mathbf v):=\boldsymbol\xi_\mu^\top \mathbf v+\mathbf b_\mu,\quad m(\mathbf v):=\max_\mu s_\mu(\mathbf v), \quad \mathbf f:=\text{softmax}(\beta \mathbf s)
\end{align}
Define the basin of attraction around the winning softmax logit $k$ by the margin $\gamma>0$:
\begin{align}
    \mathcal B_k(\gamma)=\{\mathbf v: s_k(\mathbf v)-\max_{j\ne k} s_j(\mathbf v)\ge\gamma\}
\end{align}
Let $T_I$ be the first time $t$ such that $\mathbf v(t)\in \cup_k \mathcal B_k(\gamma)$. Defining the softmax component of the energy function \eqref{eq:energy-in-visible-neurons} as
\begin{align*}
    \text{LSE}_\beta(\mathbf s)=\frac1\beta\log\sum_{\mu=1}^{N_h}e^{\beta s_\mu}
\end{align*}
then for every $\mathbf v$, we can bound the LSE as
\begin{align}
    m(\mathbf v) \le \mathrm{LSE}_\beta(s(\mathbf v)) \le m(\mathbf v) + \frac{1}{\beta}\log N_h
\label{eq:lse-band}
\end{align}
Thus, the ``softmax slack" $\delta(\mathbf v):=\text{LSE}_\beta(s(\mathbf v))-m(\mathbf v)$ obeys $0\le\delta(\mathbf v)\le\frac1\beta \log N_h$. In the high-$\beta$ regime, there are no critical points other than the softmax basins (those within $\cup_k \mathcal B_k(\gamma)$ for any reasonable $\gamma>\epsilon>0$). To reduce $\delta$ from its initial value to the cusp of one of the basins requires dissipating at most
\begin{align}
    \Delta E_\text{softmax}\le\frac{1}{\beta}\log N_h
\end{align}
$\frac{\partial E}{\partial v_i}=-\tau_v\dot v_i$, and outside winning basins $\tau_v\dot v_i\sim1$, so the squared magnitude of the gradient grows at least linearly in $N_v$:
\begin{align}
    \|\nabla E(\mathbf v)\|^2 = \sum_{i=1}^{N_v} \left(\frac{\partial E}{\partial v_i}\right)^2 \ge c N_v
\end{align}
for some $c>0$ independent of $N_v$ and $N_h$ for all $\mathbf v$ in the trajectory outside a winning basin. Therefore, the energy dissipation rate satisfies
\begin{align}
    -\dot E(t)
    = \frac{1}{\tau_v}\|\nabla E(\mathbf v(t))\|^2
    \ge \frac{c}{\tau_v}N_v
\end{align}

Under assumptions \textbf{(A1)--(A2)}, the visible state $\mathbf v$ remains in a bounded box, so the quadratic part of the energy contributes at most $\mathcal O(N_v)$ to the energy difference between any two points on the trajectory. Since the energy dissipation rate during $T_I$ scales proportionally to $N_v$, the quadratic component of the energy contribution is dissipated in constant time. The only nontrivial $N_h$ dependence is due to the softmax slack. Together with the bound on $\Delta E_\text{softmax}$, the total time this phase takes is characteristically
\begin{align}
    T_I=\mathcal O\left(\frac{\tau_v}{\beta} \frac{\log N_h}{N_v}\right)
\end{align}

\subsubsection{$T_{II}$: Contractive convergence within a winning basin}
Find a basin $\mathcal B_k(\gamma)$ that is entered at $t_\text{in}=T_I$. We will now show local strong convexity within this basin, allowing us to invoke the PL inequality and find exponential convergence within the basin.  Define $G:=\text{Diag}(\mathbf f)-\mathbf f\mathbf f^\top$.
First, consider that the non-winning softmax mass is $1-f_k$, which is
\begin{align}
    1-f_k =\sum_{j\ne k}f_j\le (N_h-1)e^{-\beta\gamma}
\end{align}
Additionally, since $\|\mathbf f\|^2=f_k^2+\sum_{j\ne k}f_j^2\ge f_k^2$ and $0\le f_k\le 1$,
\begin{align}
    \lambda_\text{max}(G(\mathbf f))\le\text{tr}(G(\mathbf f))=1-\|\mathbf f\|^2\le 1-f_k^2 \le 2(1-f_k)\le 2(N_h-1)e^{-\beta\gamma}
\end{align}
Hence, with $S^2=\max_\mu\|\boldsymbol\xi_\mu\|^2$,
\begin{align}
    \lambda_\text{max}(\boldsymbol\xi^\top G(\mathbf f)\boldsymbol\xi)\le S^2\lambda_\text{max}(G(\mathbf f))\le 2S^2(N_h-1)e^{-\beta\gamma}
\end{align}
This gives a bound on the largest eigenvalue of $G(\mathbf f)$ in a way that incorporates the softmax beta. 

Now, we can show local strong convexity in the winning basin:
\begin{align}
    \nabla^2 E(\mathbf v)=I-\beta\boldsymbol\xi^\top G(\mathbf f)\boldsymbol\xi \succeq (1-\beta 2S^2(N_h-1)e^{-\beta\gamma})I \equiv \alpha(\beta, \gamma)I
\end{align}
for all $\mathbf v\in\mathcal B_k(\gamma)$. Particularly, if 
\begin{align}
    e^{-\beta\gamma}(N_h-1)\le\frac{1}{4\beta S^2}
\end{align}
then $\alpha(\beta, \gamma)\ge\frac12$, independent of $N_h$, $N_v$. Note that this is always possible: if the softmax is not peaked enough to make this inequality true, simply keep moving in trajectory ``Phase I" for a little longer until the margin $\gamma$ grows slightly larger such that the condition holds true. This strong convexity within $\mathcal B_k(\gamma)$ implies the PL inequality
\begin{align}
    \frac12\|\nabla E(\mathbf v)\|^2\ge\alpha(\beta, \gamma)(E(\mathbf v)-E^\star),\quad \forall \mathbf v\in\mathcal B_k(\gamma)
\end{align}
Therefore, along the trajectory within the basin for times $t\ge t_\text{in}$,
\begin{align}
    \frac{d}{dt}\big(E(\mathbf v(t)) - E^\star\big)
    = -\frac{1}{\tau_v}\|\nabla E(\mathbf v(t))\|^2
    \le -\frac{2\alpha(\beta,\gamma)}{\tau_v}\big(E(\mathbf v(t)) - E^\star\big)
\end{align}
Integrating,
\begin{align}
    E(\mathbf v(t)) - E^\star
    \le e^{-\frac{2\alpha(\beta,\gamma)}{\tau_v}(t-t_\text{in})}\,\big(E(\mathbf v(t_\text{in})) - E^\star\big)
\end{align}
Impose a relative-to-initial convergence criteria:
\[
E(\mathbf v(t)) - E^\star \le \epsilon\,\big(E(\mathbf v(0)) - E^\star\big),\quad \epsilon\in(0,1)
\]
Since $E$ is non-increasing along the trajectory, $E(\mathbf v(t_\text{in})) - E^\star \le E(\mathbf v(0)) - E^\star$, so it suffices that
\[
e^{-\frac{2\alpha(\beta,\gamma)}{\tau_v}(t-t_\text{in})} \le \epsilon
\]
Hence the in-basin time satisfies
\begin{align}
    T_{II} \le \frac{\tau_v}{2\alpha(\beta,\gamma)}\log\frac{1}{\epsilon} = \mathcal{O}\!\left(\tau_v\log\frac{1}{\epsilon}\right)
\end{align}
which is size-free of $N_h$ and $N_v$.

\subsubsection{Combined bound}
Altogether, in the high-$\beta$ regime, to reach a relative-to-initial tolerance of
\begin{align}
E(\mathbf v(t)) - E^\star \le \epsilon\,\big(E(\mathbf v(0)) - E^\star\big)
\end{align}
the combined convergence time satisfies
\begin{align}
    T(\epsilon) = \underbrace{\mathcal O\left(\frac{\tau_v}{\beta} \frac{\log N_h}{N_v}\right)}_{\text{winner selection }(T_I)} \quad\;\;+ \underbrace{\mathcal{O}\!\left(\tau_v\log\frac{1}{\epsilon}\right)}_{\text{convergence within basin }(T_{II})}
\end{align}
For fixed $\epsilon$, $\beta$, and $\tau_v$, $T_{II}$ is independent of $N_v$ and $N_h$, while $T_I$ carries all the model-size dependence. The dependence of the convergence time on $N_h$ and $N_v$ in the high-$\beta$ regime is
\begin{align}
    T(\epsilon) = \mathcal O\!\left(\frac{\tau_v}{\beta}\frac{\log N_h}{N_v}\right).
\end{align}
The convergence time is at most logarithmic in the number of hidden neurons $N_h$, and actually decreases as $1/N_v$ in the number of visible neurons.

\subsection{Limitations} Our analysis assumes that the timescales of the crossbar array are much faster than the fastest neuronal timescales. In practice, as the crossbar array gets bigger, it may contribute to the time scales of the entire system, since wires have non-zero capacitances. Once the size of the crossbar array reaches the point when it significantly modifies the time scales of the neurons, our analysis and the scaling argument becomes invalid. For this reason, one cannot scale this design to infinitely large sizes. Analyzing that boundary is outside the scope of our paper, because it is dependent on fabrication and design parameters,  which is a different level of abstraction than our present paper.

\section{Design invariance under voltage scaling}\label{appendix:voltage-scaling}
Given hardware constraints of $G_\text{max}$, $C_c$, and $C_r$, we can still implement models with  arbitrarily large weights. Convergence bounds rely on the weight matrix constraints, which can be made feasible by global normalization at the hardware level, keeping the effective model weights unchanged. Consider the scaling factor for any non-negative $\boldsymbol\xi$:
\begin{align}
    \kappa=\min\left\{ 1,\frac{G_\text{max}}{\max_{\mu, i}\xi_{\mu i}}, \frac{C_c}{\max_i\sum_\mu \xi_{\mu i}}, \frac{C_r}{\max_\mu\sum_i \xi_{\mu i}} \right\}
\end{align}
Set $\tilde{\boldsymbol\xi}=\kappa\boldsymbol\xi$. Then, $\tilde{\boldsymbol\xi}$ satisfies all the hardware constraints of assumption \textbf{(A1)}:
\begin{align}
    0\le\tilde\xi_{\mu i}\le G_\text{max},\quad \sum_i\tilde \xi_{\mu i}\le C_r~~\forall \mu,\quad \sum_\mu\tilde\xi_{\mu i}\le C_c~~\forall i
\end{align}
So any $\boldsymbol\xi$ matrix can be mapped onto budgets with one scalar $\kappa$. Consider the pre-softmax arguments for the hidden neurons: if we scale weights $\boldsymbol\xi \to \tilde{\boldsymbol\xi}=\kappa\boldsymbol\xi$, rescale the voltage unit $\mathbf v\to\tilde{\mathbf v}=\kappa \mathbf v$ and biases $\mathbf b\to\tilde{\mathbf b}=\kappa^2 \mathbf b$ and set $\tilde \beta=\beta/\kappa^2$, then
\begin{align}
    \tilde\beta(\tilde{\boldsymbol\xi_\mu}^\top \tilde{\mathbf v} +\tilde{\mathbf b})=\beta(\boldsymbol\xi_{\mu}^\top \mathbf v+\mathbf b)
\end{align}
so the softmax outputs $f$ and the system's attractors are unchanged. The visible ODE $\tau_v\dot {\mathbf v}=\boldsymbol\xi^\top f(\mathbf v)-\mathbf v$ is preserved up to units, as the $\kappa$ terms can be absorbed into the gain of \texttt{U2} and \texttt{U3} without affecting the convergence time bounds.

\section{Scaling of energy consumption}\label{appendix:energy-scaling}
The energy consumption of DenseAM circuits can be broken up into two parts: the energy dissipated by the weights as a result of Ohm's Law, and the energy from engineering overhead found in amplifiers and active circuitry. The energy dissipated by the weights in the crossbar array can be expressed as the integral of the power dissipated by each resistor of resistance $R_{\mu i}$ from time 0 until convergence at $T_{\text{conv}}$.

\paragraph{Energy consumption of weights.}
Let the neuron output voltages be proportional to activations: $u_i=\kappa g_i$ and $w_\mu=\kappa f_\mu$, where $\kappa$ is a fixed voltage scale. We assume rail-bounded outputs $|u_i|\le\kappa$ and $|w_\mu|\le\kappa$ (by Appendix~\ref{appendix:voltage-scaling}, global rescaling of $\boldsymbol\xi$, voltages, and $\beta$ preserves the DenseAM dynamics, so this choice of $\kappa$ does not affect behavior.)
The instantaneous power in the resistive crossbar is:
\begin{align}
    P_\text{weights}(t)=\sum_{i,\mu}\xi_{\mu i}(u_i-w_\mu)^2
\end{align}
Using the row/column conductance budgets $\sum_\mu\xi_{\mu i}\le C_c$ and $\sum_i\xi_{\mu i}\le C_r$ (Appendix~\ref{appendix:inference-scaling}) and the inequality $(a-b)^2\le2a^2+2b^2$,
\begin{align}
        P_\text{weights}(t)&\le 2\left(\sum_{i,\mu}\xi_{\mu i}u_i^2+\sum_{i,\mu}\xi_{\mu i}w_\mu^2\right)\\
    &=2\left( \sum_i u_i^2\left(\sum_\mu\xi_{\mu i}\right)+\sum_\mu w_\mu^2\left(\sum_i \xi_{\mu i}\right) \right)\\
    &\leq2\left(C_c\sum_i u_i^2+C_r\sum_\mu w_\mu^2\right)
\end{align}
If the hidden layer uses a softmax activation, then $\sum_\mu f_\mu^2\le 1$ and so $\sum_\mu w_\mu^2\le \kappa^2$; and rail bounds give $\sum_i u_i^2\le N_v\kappa^2$. Therefore,
\begin{align}
    P_\text{weights}(t)\le 2\kappa^2(C_c N_v + C_r)=\mathcal O(N_v)
\end{align}
Therefore, a system taking time $T^\text{conv}$ to converge results in an energy consumption of
\begin{align}
    E_\text{weights}=\int_0^T P_\text{weights}(t)dt \le 2\kappa^2(C_cN_v+C_r)T^\text{conv}
\end{align}
According to the convergence time bounds of \autoref{appendix:inference-scaling}, $T^\text{conv}=\mathcal O(\tau_v)$. Thus,  $E_\text{weights}=\mathcal O(N_v)$, as a function of system size. 

\paragraph{Energy consumption of capacitors.} Let each neuron node voltage be bounded by hardware limits $|u_i(t)|,~|w_\mu(t)|\le\kappa$. Charging a capacitor of capacitance $C$ from a supply through a resistive path draws $CV^2$ from the power supply. The number of times each capacitor charges is finite because the Lyapunov energy of the DenseAM forbids limit cycles. This means the total supply energy per node can be bounded by a constant. Therefore, the total energy needed to (re)charge all neuron capacitors is bounded by
\begin{align}    E_\text{capacitors}\le \mathcal O(1) \cdot \kappa^2\left(\sum_{i=1}^{N_v} C_i^{(v)} + \sum_{\mu=1}^{N_h} C_\mu^{(h)}\right)=\mathcal O(N_v+N_h)
\end{align}

\paragraph{Energy consumption of amplifiers, bias, control, and overhead.} Per neuron, the energy expenditure to amplifier inefficiency, bias terms, and general overhead do not depend on system size. For a runtime of duration $T^\text{conv}$, the energy consumption of these elements in the entire network scales as
\begin{align}
    E_\text{other}=\mathcal O((N_v+N_h)T^\text{conv})
\end{align}

\paragraph{Combined energy consumption.} All together, the total energy consumption can be written as
\begin{align}
    E_\text{total}= \mathcal O(N_v+N_h) 
\end{align}

\section{Model Specifications and Details}
\autoref{tab:xor-model-spec}, \autoref{tab:hamming-model-spec}, and \autoref{tab:parity-model-spec} summarize the model design for the XOR, Hamming (7,4), and parity DenseAM models.

\begin{table}
\caption{XOR model specification}
\centering
\begin{tabularx}{\linewidth}{@{}lX@{}}
\toprule
Visible neurons $v_i$ & $N_v=3$ (inputs $v_1,v_2$ clamped to \{0,1\}; output $v_3$ free) \\
Hidden neurons $h_\mu$  & $N_h=4$ (one per truth-table row) \\
Visible activation and Lagrangian     & Identity: $g_i=v_i$, \ $\mathcal{L}_v = \frac{1}{2}\sum_{i=1}^{N_v} v_i^2$ \\
Hidden activation and Lagrangian      & Softmax: $f_\mu=\mathrm{softmax}(\beta h_\mu)$, \ $\mathcal{L}_h = \frac{1}{\beta}\log\Big(\sum\limits_{\mu=1}^{N_h} e^{\beta h_\mu}\Big)$ \\
Visible biases          & $a_i=0$ \\
Hidden biases           & $b_\mu=-\tfrac12\sum_{i=1}^{N_v}\xi_{\mu i}^2$ \\
Weights $\boldsymbol{\xi}$           & $\boldsymbol{\xi}\in\{0,1\}^{4\times3}$, rows encode memories:
$\displaystyle
\boldsymbol{\xi}=\begin{bmatrix}
0&0&0\\
0&1&1\\
1&0&1\\
1&1&0\\
\end{bmatrix}
$ \\
Inference protocol      & Clamp $(v_1,v_2)$ to input values; read out $v_3$ at convergence \\
\bottomrule
\end{tabularx}
\label{tab:xor-model-spec}
\end{table}

\begin{table}
\caption{Hamming (7,4) model specification}
\centering
\begin{tabularx}{\linewidth}{@{}lX@{}}
\toprule
Visible neurons ($N_v$) & 7 (codeword bits)\\
Hidden neurons ($N_h$)  & 16 (one per valid codeword)\\
Visible activation      & Identity: $g_i=v_i$\\
Hidden activation       & Softmax over $\mu\in\{1,\dots,16\}$ with temperature $\beta$\\
Visible biases          & $a_i=0$\\
Hidden biases           & $b_\mu=-\tfrac12\sum_{i=1}^{N_v}\xi_{\mu i}^2$\\
Weights $\xi$           & $\boldsymbol \xi\in\{0,1\}^{16\times7}$, each row is a valid Hamming(7,4) codeword\\
Inference protocol      & Initialize visible neurons to corrupted 7-bit input codeword; let all visible and hidden neurons evolve; converged visible neurons give the corrected codeword\\
\bottomrule
\end{tabularx}
\label{tab:hamming-model-spec}
\end{table}

\begin{table}
\centering
\caption{8-bit parity model specification}
\begin{tabularx}{\linewidth}{@{}lX@{}}
\toprule
Visible neurons $v_i$ & $N_v=16$ (dimension of embedding $D$)\\
Hidden neurons (energy attention) $h_A^\text{attn}$ & $N_h^\text{attn}=8$ (context length $L$)\\
Hidden neurons (Hopfield network) $h_\mu^\text{hopf}$ & $N_h^\text{hopf}=16$ (Hopfield network memories $M$)\\
Hidden neurons (total) & $N_h =24$ ($L+M$)\\
\addlinespace[0.5em] 
Visible activation      & Identity: $g_i=v_i$ \\
Hidden activation (energy attention) & Softmax: $f^\text{attn}_A = \text{softmax}(\beta\mathbf h^\text{attn})_A$ for $A=1,\dots,L$\\
Hidden activation (Hopfield network) & ReLU: $f^\text{hopf}_\mu=\max{(h^\text{hopf}_\mu, 0)}$ for $\mu=1,\dots,M$\\
Weights (energy attention) &  $\boldsymbol\xi^\text{attn}\in\mathbb R^{L\times D}$, where $\boldsymbol\xi^\text{attn}_{A}$ is embedded $A$'th context token\\
Weights (Hopfield network) & $\boldsymbol\xi^\text{hopf}\in\mathbb R^{M\times D}$, static after training\\
Inference protocol         & Embed $L$ context tokens to obtain $\boldsymbol\xi^\text{attn}$. Let visible neurons evolve until convergence\\
\bottomrule
\end{tabularx}
\label{tab:parity-model-spec}
\end{table}

\subsection{Bit string energy transformer implementation}\label{appendix:parity-training}
As described in \autoref{tab:parity-model-spec}, our trained model uses an embedding matrix of $2\times D=32$ parameters, the Hopfield network with $D\times M=256$ parameters, an additional $D\times2=32$ parameter matrix to decode embeddings to logits, a total of $D+L+M=40$ neuron bias terms, and $2$ biases for the linear decoder. This is a total of $362$ parameters. 

In training and inference we use time constants $\tau_v=0.1$ and $\tau_h=0.01$. We train with Euler steps of 1e-3, and test with Euler steps of 1e-4 for a time horizon of $T=1$ second. Jax's automatic differentiation was used to implement backpropagation through time. We encourage the model to reach fixed points by penalizing $\dot v$ at time T. This yields models that are more robust to hardware imperfection due to the intrinsic stability of attractor points. The convergence to an attractor also means the inference remains stable to mismatch and delay in timing during readout. 

\section{Hardware analysis}
\subsection{Hardware speed analysis}
As discussed in \autoref{sec:inference-time-scaling}, the convergence time of analog DenseAMs is governed not by system size, but rather primarily by the timescales of the dynamics in hardware. These timescales are set by the time constants $\tau_v$ and $\tau_h$. The smaller these time constants, the faster the dynamics move, and the faster the system converges. In this section, we derive bounds on the minimum time constant $\min\{\tau_v, \tau_h\}$ of the DenseAM, which is limited by the constraints of active components like amplifiers. 

The maximum speed of neuronal dynamics is limited by the ability of active stages (op-amps/buffers) to track changing signals. If the input slope to an active stage exceeds its slew rate (SR), the output distorts; if the signal spectrum approaches or exceeds the stage’s closed-loop bandwidth, attenuation and phase lag appear. Here, we derive lower bounds on the time constants $\tau_v,\tau_h$ imposed by (i) finite gain–bandwidth product (GBW) and (ii) finite SR of the three active stages in the neuron design (\autoref{appendix:neuron-design}). Without loss of generality we will express the derivation for the hidden neurons, with the derivations for visible neurons following by symmetry. Throughout, define the following: 
\begin{itemize}
    \item State swing: $|v_i(t)|\le A_v$, so that $|\dot v_i|\lesssim A_v/\tau$. Similarly, $|h_\mu(t)|\le A_h$, so that $|\dot h_\mu|\lesssim A_h/\tau$.
    \item Activation swing: Visible activation $g(\cdot)$ is Lipschitz with slope bound $L_g=\sup_x|g'(x)|$. Then, $|\dot g_i|\le L_g |\dot v_i|\le L_g A_v/\tau$. Similarly, hidden activation $f(\cdot)$ is Lipschitz with slope bounded by $L_f=\sup_x|f'(x)|$. Then, $|\dot f_\mu|\le L_f|\dot h_\mu|\le L_f A_h / \tau$.
    \item Weights $\boldsymbol\xi\ge0$. Hardware normalization gives per-row/column conductivity budgets, so the self-term gain for hidden neuron $\mu$ is $A_{\text{self},\mu}=\sum_i\xi_{\mu i}=\mathcal O(1)$. 
\end{itemize}
We will derive three independent lower bounds and then take the max:
\begin{align}
    \tau_\text{min}\ge\max\{\underbrace{\tau_\text{GBW}}_\text{tracking small signals}, \underbrace{\tau_\text{SR}}_\text{edge/large-signals}, \underbrace{\tau_{I-\text{limit}}}_{\text{output current}}\}
\end{align}

\subsubsection{Gain-bandwidth product bound}
For a single-pole op-amp with gain-bandwidth product $\text{GBW}$ in a closed-loop configuration with loop gain $A_\text{CL}$, the $-3$db bandwidth is $f_c\approx\text{GBW}/A_\text{CL}$. In order for the neuron to faithfully track with a time constant $\tau$, we require $f_c\gtrsim1/(2\pi\tau)$ for every stage in the signal path.     Closed-loop gains for each of the op-amps are: $A_\text{CL}(U1)=1$ because it is a unity-gain buffer, $A_\text{CL}(U2)=A_\text{self}$ because it needs to realize the self term gain, and $A_\text{CL}(U3)\approx 1$ because it is a unity-gain summer.
Assuming the same op-amp design for \texttt{U1}, \texttt{U2}, and \texttt{U3}, and taking the worst case, 
\begin{align}
    \tau_\text{GBW}=\frac{\max(1, A_\text{self})}{2\pi\text{GBW}}
    \label{eq:tau_gbw}
\end{align}

\subsubsection{Slew rate bound}
The slew-rate limits cap the maximum output slope of each op-amp stage:
\begin{itemize}
    \item \textbf{\texttt{U1}: activation buffer.} $|\dot f_\mu|\le L_fA_h/\tau$, which gives $\tau\ge(L_f A_h)/\text{SR}_\text{U1}$.
    \item \textbf{\texttt{U2}: self-term.} $s_\mu=A_\text{self} f_\mu$, so $|\dot s_\mu|=A_\text{self}|\dot f_\mu|\le (A_\text{self}L_fA_h)/\tau$, which gives $\tau\ge (A_\text{self}L_fA_h)/\text{SR}_\text{U2}$.
    \item \textbf{\texttt{U3}: internal state drive.} The time-varying portion of the RC circuit drive $d_\mu$ is a linear combination of $f_\mu$ and $g_i$, with coefficients that have a maximum magnitude of $A_\text{self}$. Using the bounds on the slopes of those inputs, we get the following bound on $|\dot d_\mu|$ and subsequently the time constant bound:
    \begin{align}
        |\dot d_\mu|\lesssim\frac{A_\text{self}}{\tau}\max\{L_fA_h,L_gA_v\} \quad\Rightarrow\quad \tau\ge\frac{A_\text{self}\max(L_fA_h,L_gA_v)}{\text{SR}_\text{U3}}
    \end{align}
\end{itemize}
All together, the combined constraint is
\begin{align}
    \tau_\text{SR}=\max\left\{ \frac{L_fA_h}{\text{SR}_\text{U1}},\frac{A_\text{self}L_fA_h}{\text{SR}_\text{U2}}, \frac{A_\text{self}\max(L_fA_h,L_gA_v)}{\text{SR}_\text{U3}} \right\}
    \label{eq:tau_sr}
\end{align}

\subsubsection{Current / headroom limit}
\texttt{U3} must provide the current through $R_2$ to charge $C_1$. The RC circuit dynamics dictate $R_2C_1\dot h_\mu=-h_\mu+d_\mu$, so the instantaneous current needed by \texttt{U3} is
\begin{align}
    I_\text{U3,out}=\frac{d_\mu-h_\mu}{R_2}=C_1\dot h_\mu
\end{align}
We must respect $|I_\text{U3,out}|\le I_\text{max,U3}$. With $|\dot h_\mu|\lesssim A_h/\tau$, 
\begin{align}
    \tau_\text{I-limit}\ge\frac{C_1A_h}{I_\text{max,U3}}
    \label{eq:tau-i-limit}
\end{align}

\subsubsection{Combined bound on minimum time constant}
Taken together, the minimum time constant must satisfy the bounds \eqref{eq:tau_gbw}, \eqref{eq:tau_sr}, and \eqref{eq:tau-i-limit}:
\begin{align}
    \tau_\text{min}\ge\max\{\tau_\text{GBW}, \tau_\text{SR}, \tau_\text{I-limit}\}
    \label{eq:combined-min-tau-bound}
\end{align}

\subsection{Estimates of inference times with existing hardware}\label{appendix:existing-hardware}
Under standard assumptions for DenseAMs (symmetric couplings and monotone activations), the Lyapunov energy decreases monotonically and the dynamics converge without oscillations. The settling time is therefore on the order of a few multiples of the largest neuronal time constant, which we bound by amplifier non-idealities. In this section we take some representative examples of op-amps from literature and estimate the inference speeds from reasonable and representative design parameters.

\begin{table}
\centering
\small
\caption{Estimated neuron time constants and conservative convergence times with $A_v=A_h=1$\,V, $L_g=1$, $A_\text{self}=1$ for representative amplifiers in literature. 
GBW bound $\tau_{\GBW}=\frac{1}{2\pi\,\GBW}$; SR bound $\tau_{\SR}=\frac{L_g A_v}{\SR}$ (visible path).
Overall $\tau_{\min}=\max\{\tau_{\GBW},\tau_{\SR}\}$; we report $T_{\text{conv}}=10\,\tau_{\min}$.}

\label{tab:ota_1V}
\begin{tabular}{@{}lrrrrc@{}}
\toprule
\textbf{CMOS Amplifier (ref.)} & \textbf{SR (V/$\mu$s)} & \textbf{GBW (MHz)} & \textbf{$\tau_{\SR}$ (ns)} & \textbf{$\tau_{\GBW}$ (ns)} & \textbf{$T_{\text{conv}}$ (ns)} \\
\midrule
\citet{perez2012performance}                & 84.50  & 321.50 & \cellcolor{lightgray}{11.83} & 0.50  & \textbf{118.34} \\
\citet{assaad2009recycling}         & 94.10  & 134.20 & \cellcolor{lightgray}{10.63} & 1.19  & \textbf{106.27} \\
\citet{yen2020high}                 & 202.00 & 10.70  & 4.95  & \cellcolor{lightgray}{14.87} & \textbf{148.74} \\
\citet{naderi2018operational}       & 1250.00& 3600.00& \cellcolor{lightgray}{0.80}  & 0.04  & \textbf{8.00}   \\
\citet{schlogl2007design} & 1650.00 & 2510.00 & \cellcolor{lightgray}{0.61} & 0.06 & \textbf{6.06}\\
\bottomrule
\end{tabular}

\vspace{4pt}
\footnotesize\emph{Notes.} (i) $\tau_{\SR}$ values assume the visible path dominates the summer’s SR (low/moderate-$\beta$). 
If softmax dominates at U3 in the high-$\beta$ regime, multiply SR-limited values by $\kappa = ( \beta/2 )\,(A_h/A_v)$ (with $A_h=A_v=1$\,V, simply $\beta/2$). 
(ii) The current-limit bound $\tau_{I\text{-limit}} = C A_v / I_{\max}$ is typically $\ll$ all reported values for $C\!\sim\!50$\,fF and $I_{\max}\!\sim\!$mA, so it is omitted from the table but must still be respected in circuit sizing.
\end{table}

\paragraph{Minimum time constant.}
For illustration purposes, we choose three reasonable hardware constraints:
\begin{itemize}
    \item \textbf{Activation slopes.} Take the slope of the visible activation to be $L_g=1$, such as would occur in a identity visible neuron activation. Take the worst-case (maximum) slope of the hidden activation to be according to the softmax with fixed $\beta$, whose Jacobian is $\beta G(\mathbf f)$ with $\|G(\mathbf f)\|_2\le \frac12$, so a safe global bound is $L_f\le\frac\beta2$. 
    \item \textbf{Signal swing.} Use the voltage scaling invariance (see \autoref{appendix:voltage-scaling}) to rescale $\mathbf v$, $\boldsymbol\xi$, and $\beta$ together to pick a swing that is slew-rate friendly but well above component noise limits. Take both $A_v=A_h=1V$.
    \item \textbf{Self-term gain.} With row/column budgets, use $A_\text{self}$ as a worst-case bound.
\end{itemize}
With those choices, the three lower bounds per neuron are:
\begin{enumerate}
    \item \textbf{GBW Bound:} $\tau_\text{GBW}=\frac{\max(1, A_\text{self})}{2\pi\text{GBW}}=\frac{1}{2\pi\text{GBW}}$.
    \item \textbf{SR Bound:} The U1/U2 path give $\tau_\text{SR,vis}=\frac{L_gA_v}{\text{SR}}=\frac{1}{\text{SR}}$ µs. In the U3 (summer) path, equation~\eqref{eq:tau_sr} has two cases.
    In the low-$\beta$ regime where $L_gA_v\ge L_fA_h$, then U3 bound reduces to $1/\text{SR}$ µs. In the high-$\beta$ regime where $L_fA_h=\beta/2$ dominates, scale the slew-rate limited bound by $\beta/2$. 
    \item \textbf{Output Current Bound:} In practice, this bound generally does not limit the op amp choice: even with a large capacitor $C=50$ fF, $A_v=1$V, $I_\text{max}=2$mA, $\tau_\text{I-limit}\approx0.025$ns, which is negligible compared to the bounds from SR and GBW. 
\end{enumerate}
To quantify realistic inference speeds, Table~\ref{tab:ota_1V} lists representative CMOS operational transconductance amplifiers (OTAs)\footnote{Many high-speed CMOS “op-amps” are reported as OTAs (transconductors). In our neuron, these OTA cores operate in closed-loop (unity/non-inverting) configurations, so the literature SR and GBW directly constrain $\tau$ via Eqs.~\eqref{eq:tau_gbw}–\eqref{eq:tau_sr}.} drawn from recent literature, together with their corresponding lower bounds on neuronal time constants under the GBW and slew‐rate limits.
Even using conservative assumptions with existing amplifier designs, the analysis shows that modern high‐speed OTAs can achieve sub–10 ns neuronal convergence times—corresponding to inference rates in the hundreds of megahertz.

\section{Connection between analog and canonical Energy Transformer}\label{appendix:et-energy-function}
In this section we show that in the adiabatic limit, our Analog Energy Transformer (Analog ET) reduces to the canonical Energy Transformer. Begin with the dynamics for the Analog Energy Transformer implemented by our circuit designs.
\begin{align}
    \tau_v \dot{\mathbf{v}} 
        &= -\frac{\partial E}{\partial \mathbf{v}}
           = \left(\boldsymbol{\xi}^{\text{attn}}\right)^\top \mathbf{f}^{\text{attn}}
             + \left(\boldsymbol{\xi}^{\text{hopf}}\right)^\top \mathbf{f}^{\text{hopf}} + \mathbf a - \mathbf v\\
    \tau_h \dot{\mathbf{h}}^{\text{attn}} 
        &= -\frac{\partial E}{\partial \mathbf{f}^{\text{attn}}}
           = \boldsymbol{\xi}^{\text{attn}} \mathbf{v} 
             + \mathbf{b} - \mathbf{h}^{\text{attn}}\\
    \tau_h \dot{\mathbf{h}}^{\text{hopf}} 
        &= -\frac{\partial E}{\partial \mathbf{f}^{\text{hopf}}}
           = \boldsymbol{\xi}^{\text{hopf}} \mathbf{v} 
             + \mathbf{c} - \mathbf{h}^{\text{hopf}}
\end{align}
Integrating out hidden neurons in the adiabatic limit where $\tau_h\to0$, we see the relations
\begin{align}
    \mathbf h^\text{attn}(\mathbf v)&=\boldsymbol\xi^\text{attn}\mathbf v+\mathbf b\\
    \mathbf h^\text{hopf}(\mathbf v)&=\boldsymbol\xi^\text{hopf}\mathbf v+\mathbf c
\end{align}
which we can use to integrate out the hidden neuron activations as
\begin{align}
\mathbf f^\text{attn}(\mathbf v)&=\text{softmax}\left(\boldsymbol\xi^\text{attn}\mathbf v+\mathbf b\right)\\
\mathbf f^\text{hopf}(\mathbf v)&=\text{ReLU}\left(\boldsymbol\xi^\text{hopf}\mathbf v+\mathbf c\right)
\end{align}
Substituting into the visible dynamics:
\begin{align}
    \tau_v\dot{\mathbf v}=
    \left(\boldsymbol{\xi}^{\text{attn}}\right)^\top \mathbf{f}^{\text{attn}}(\mathbf v)
             + \left(\boldsymbol{\xi}^{\text{hopf}}\right)^\top \mathbf{f}^{\text{hopf}}(\mathbf v) 
             + \mathbf a - \mathbf v
\end{align}
We can ask ourselves, what scalar energy produces this ODE? We seek an energy $E_\text{eff}(\mathbf v)$ such that $\tau_v\dot{\mathbf v}=-\frac{\partial E_\text{eff}}{\partial \mathbf v}$. Equivalently, 
\begin{align}
    \nabla_\mathbf v E_\text{eff}(\mathbf v)=\mathbf v - \mathbf a -    \left(\boldsymbol{\xi}^{\text{attn}}\right)^\top \mathbf{f}^{\text{attn}}(\mathbf v) - \left(\boldsymbol{\xi}^{\text{hopf}}\right)^\top \mathbf{f}^{\text{hopf}}(\mathbf v) 
\end{align}
We can construct $E_\text{eff}(\mathbf v)$ as a sum of three pieces whose gradients match each term $E_\text{eff}(\mathbf v)=E_\text{quad}(\mathbf v)+E_\text{attn}(\mathbf v)+E_\text{hopf}(\mathbf v)$. By inspection we see that $E_\text{quad}(\mathbf v)=\frac12\|\mathbf v-\mathbf a\|^2$.

\paragraph{Attention term.} The energy function
\begin{align}
    E_\text{attn}(\mathbf v)=-\frac1\beta\log\sum_A\exp\left(\beta \left(\boldsymbol\xi_A^\text{attn} \mathbf v+b_A\right)\right)
\end{align}
satisfies our requirement. We can see that by differentiating with respect to $v_i$, we get
\begin{align}
    \frac{\partial E_\text{attn}}{\partial v_i}&=-\sum_A \text{softmax}(\boldsymbol \xi^\text{attn}\mathbf v+\mathbf b)_A\cdot\xi_{A i}^\text{attn}\\
    &=-\sum_A\xi_{A i}^\text{attn}f_A^\text{attn}(\mathbf v)
\end{align}
which yields our desired dynamics of $\nabla_\mathbf v E_\text{attn}(\mathbf v)=-\left(\boldsymbol \xi^\text{attn}\right)^\top\mathbf f^\text{attn}(\mathbf v)$.

\paragraph{Hopfield term.} A simple way to achieve the desired dynamics is with a Hopfield-type energy function
\begin{align}
    E_\text{hopf}(\mathbf v) = -\sum_\mu \frac12\left(\text{ReLU}\left(\boldsymbol\xi_\mu^\text{hopf}\mathbf v + c_\mu\right)\right)^2
\end{align}
whose derivative with respect to $v_i$ yields
\begin{align}
    \frac{\partial E_\text{hopf}}{\partial v_i}&=-\sum_\mu \text{ReLU}\left(\boldsymbol\xi_\mu^\text{hopf}\mathbf v + c_\mu\right)\cdot\xi_{\mu i}^\text{hopf}\\
    &=-\sum_\mu \xi_{\mu i}^\text{hopf}f_\mu^\text{hopf}(\mathbf v)
\end{align}
which yields our desired dynamics of $\nabla_\mathbf v E_\text{hopf}(\mathbf v)=-\left(\boldsymbol\xi^\text{hopf}\right)^\top \mathbf f^\text{hopf}(\mathbf v)$.

\paragraph{Effective energy function of analog energy transformer.} All together, the effective scalar energy over the visible state $\mathbf v$ after integrating out hidden neurons is
\begin{align}
    E_\text{eff}(\mathbf v)=\underbrace{\frac12\|\mathbf v-\mathbf a\|_2^2}_{E_\text{quad}}
    \underbrace{-\frac1\beta\log\sum_A\exp\left(\beta \left(\boldsymbol\xi_A^\text{attn} \mathbf v+b_A\right)\right)}_{E_\text{attn}}
    \underbrace{-\sum_\mu \frac12\left(\text{ReLU}\left(\boldsymbol\xi_\mu^\text{hopf}\mathbf v + c_\mu\right)\right)^2}_{E_\text{hopf}}
\end{align}
This effective energy aligns with the canonical Energy Transformer's energy function. Because our effective dynamics use hidden neurons, the energy function written in the main text reflects the contributions of the hidden neurons. When $\tau_h\ll\tau_v$, this regime converges to the behavior when the hidden neurons are integrated out. Hence, the effective expressibility and behavior of our system is equivalent to that of the original Energy Transformer. 

In our model we omit the layer normalization activation that the original Energy Transformer applies to the visible neurons. This keeps the circuit design simple, while still enabling models with high expressibility. This choice does not modify the structure of the attention or the Hopfield parts of the energy; only the self-energy of $\mathbf v$ differs. From a modeling perspective, layer normalization mainly improves conditioning and learning of deep networks rather than changing the computational primitive and expressibility. We empirically observe that the resulting models without layer normalization remain expressive enough to solve the problems we present. In principle, a layer normalization-type visible activation function could be implemented in analog hardware (e.g. by subtracting the mean voltage and normalizing by an on-chip variance estimate), but this would add distracting complications to the minimalist neuron and circuit designs we show in this paper. 

\end{document}